\pdfoutput=1

\documentclass[10pt,twocolumn,letterpaper]{article}

\usepackage{cvpr}
\usepackage{times}
\usepackage{epsfig}
\usepackage{graphicx}
\usepackage{amsmath}
\usepackage{amssymb}

\usepackage{subfig,caption}
\usepackage{svg}
\usepackage{tabls}
\usepackage{float}
\usepackage{authblk}
\usepackage{verbatim}

\newcommand\Mark[1]{\textsuperscript#1}

\DeclareMathOperator*{\argmin}{arg\,min}
\renewcommand{\vec}[1]{\mathbf{#1}}

\usepackage[breaklinks=true,bookmarks=false]{hyperref}

\cvprfinalcopy 

\def\cvprPaperID{2191} 

\begin{document}

\title{Multimodal Transfer: A Hierarchical Deep Convolutional Neural Network for Fast Artistic Style Transfer}

\author{
Xin Wang\Mark{{1,2}}, \quad Geoffrey Oxholm\Mark{2}, \quad Da Zhang\Mark{1}, \quad Yuan-Fang Wang\Mark{1} 
\\
\Mark{1}University of California, Santa Barbara,
CA
\\ 
\Mark{2}Adobe Research, San Francisco, CA
\\
{\tt\small \{xwang,dazhang,yfwang\}@cs.ucsb.edu, oxholm@adobe.com} 
}

\maketitle

\begin{abstract}

Transferring artistic styles onto everyday photographs has become an extremely popular task in both academia and industry. 
Recently, offline training has replaced on-line iterative optimization, enabling nearly real-time stylization. 
When those stylization networks are applied directly to high-resolution images, however, the style of localized regions often appears less similar to the desired artistic style. 
This is because the transfer process fails to capture small, intricate textures and maintain correct texture scales of the artworks. 
Here we propose a multimodal convolutional neural network that takes into consideration faithful representations of both color and luminance channels, and performs stylization hierarchically with multiple losses of increasing scales. 
Compared to state-of-the-art networks, our network can also perform style transfer in nearly real-time by conducting much more sophisticated training offline. 
By properly handling style and texture cues at multiple scales using several modalities, we can transfer not just large-scale, obvious style cues but also subtle, exquisite ones. 
That is, our scheme can generate results that are visually pleasing and more similar to multiple desired artistic styles with color and texture cues at multiple scales.
   
\end{abstract}

\vspace*{-2ex}
\section{Introduction}

\begin{figure*}
\vspace*{-2ex}
\begin{center}

	\setlength{\tabcolsep}{-0.5pt}
    \renewcommand{\arraystretch}{1}
	\begin{tabular}{c c c c c c}
	\subfloat[Style]{
    	\begin{tabular}{c}
    	\includegraphics[width = 1.1in]{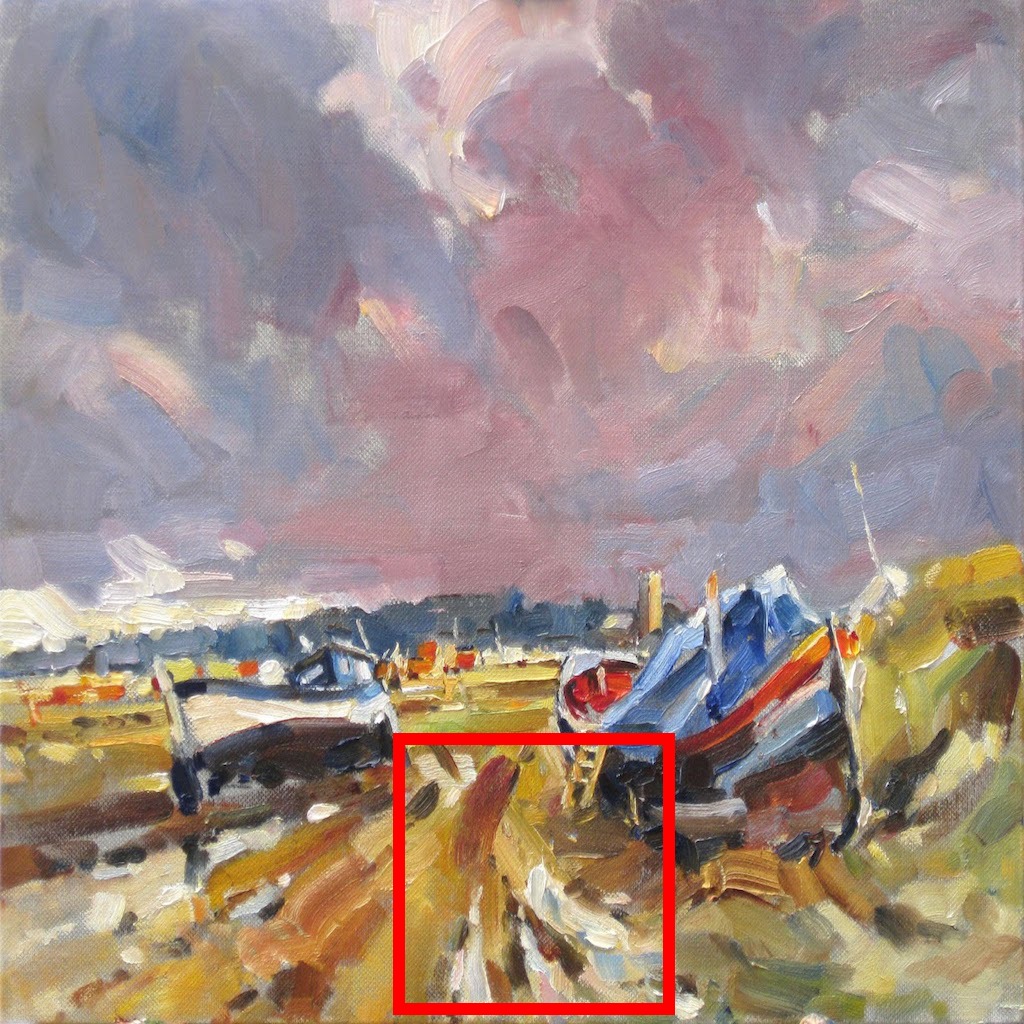} \\
    \includegraphics[width = 1.1in,height = 1.1in, trim={450 22 362 750},clip]{images/intro_figure_1/day_style.jpg}
    	\end{tabular}
    } &
	\subfloat[Gatys \textit{et al.}]{
    	\begin{tabular}{c}
    	\includegraphics[width = 1.1in]{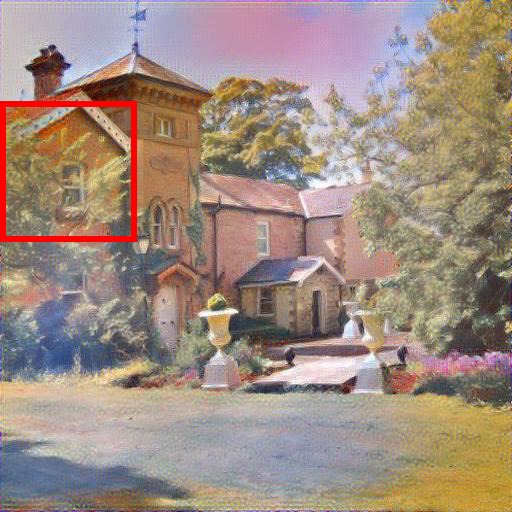} \\
    \includegraphics[width = 1.1in,height = 1.1in, trim={8 278 382 110},clip]{images/intro_figure_1/day_gatys.jpg}
    	\end{tabular}
    } &
    \subfloat[Johnson \textit{et al.}]{
    	\begin{tabular}{c}
    	\includegraphics[width = 1.1in]{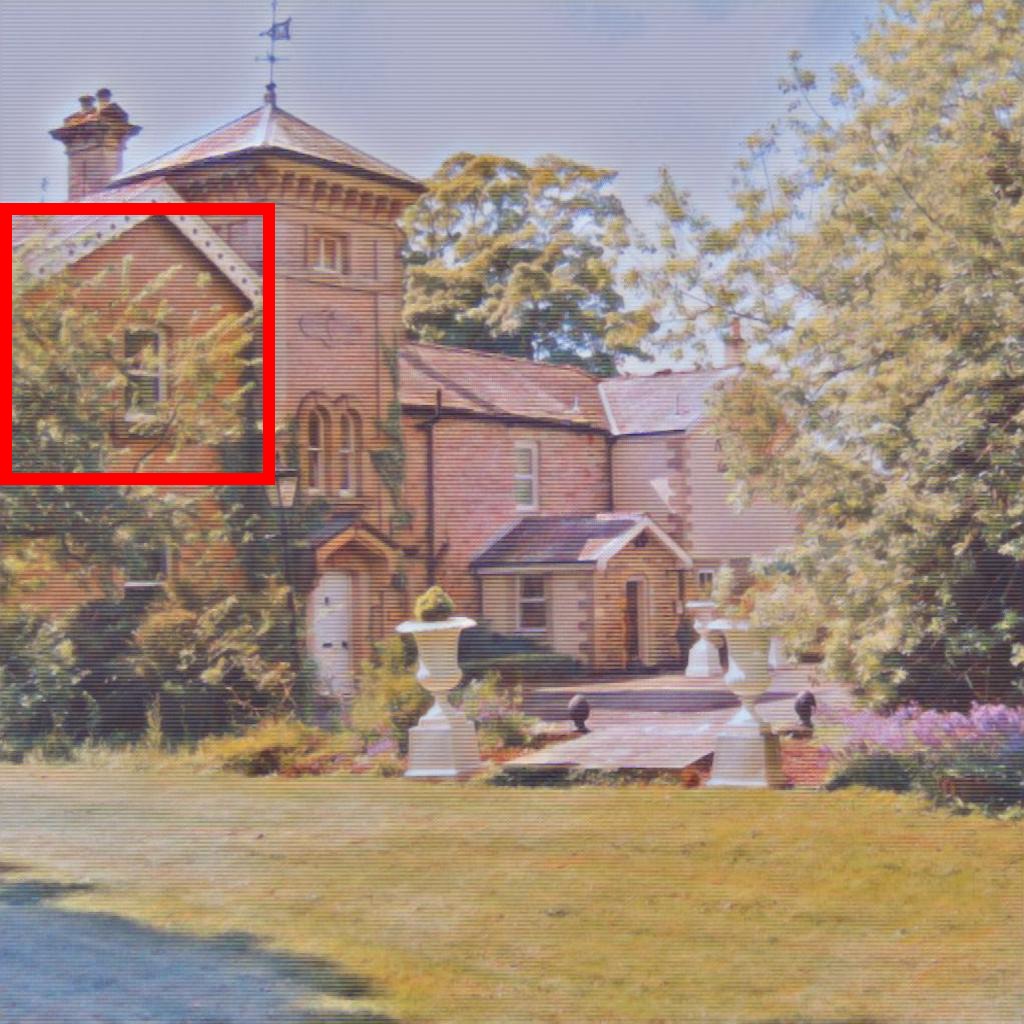} \\
    	\includegraphics[width = 1.1in,height = 1.1in, trim={12 556 764 216},clip]{images/intro_figure_1/day_johnson.jpg}
    	\end{tabular}
    } &
    \subfloat[Ulyanov \textit{et al.}]{
    	\begin{tabular}{c}
    	\includegraphics[width = 1.1in]{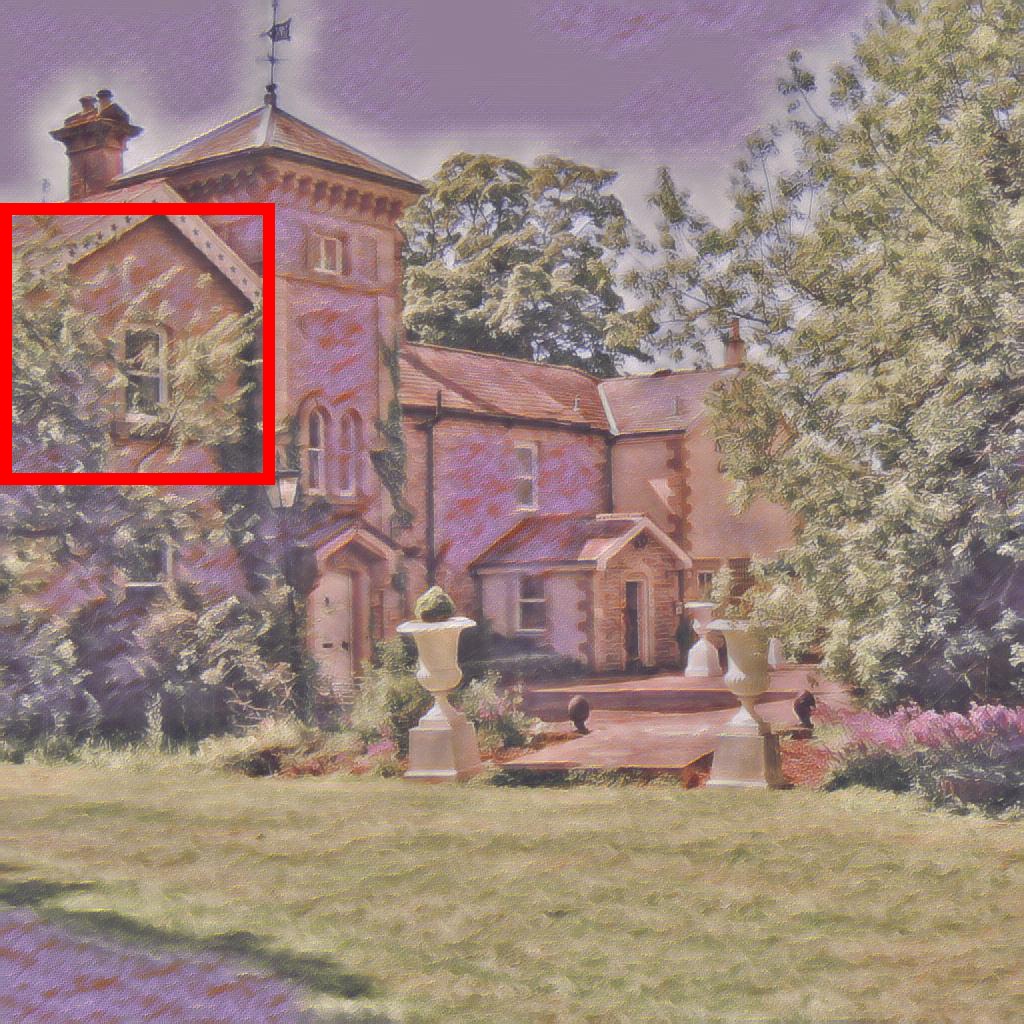} \\
    	\includegraphics[width = 1.1in,height = 1.1in, trim={12 556 764 216},clip]{images/intro_figure_1/day_pyramid.jpg}
    	\end{tabular}
    } &
    \subfloat[Ours]{
    	\begin{tabular}{c}
    	\includegraphics[width = 1.1in]{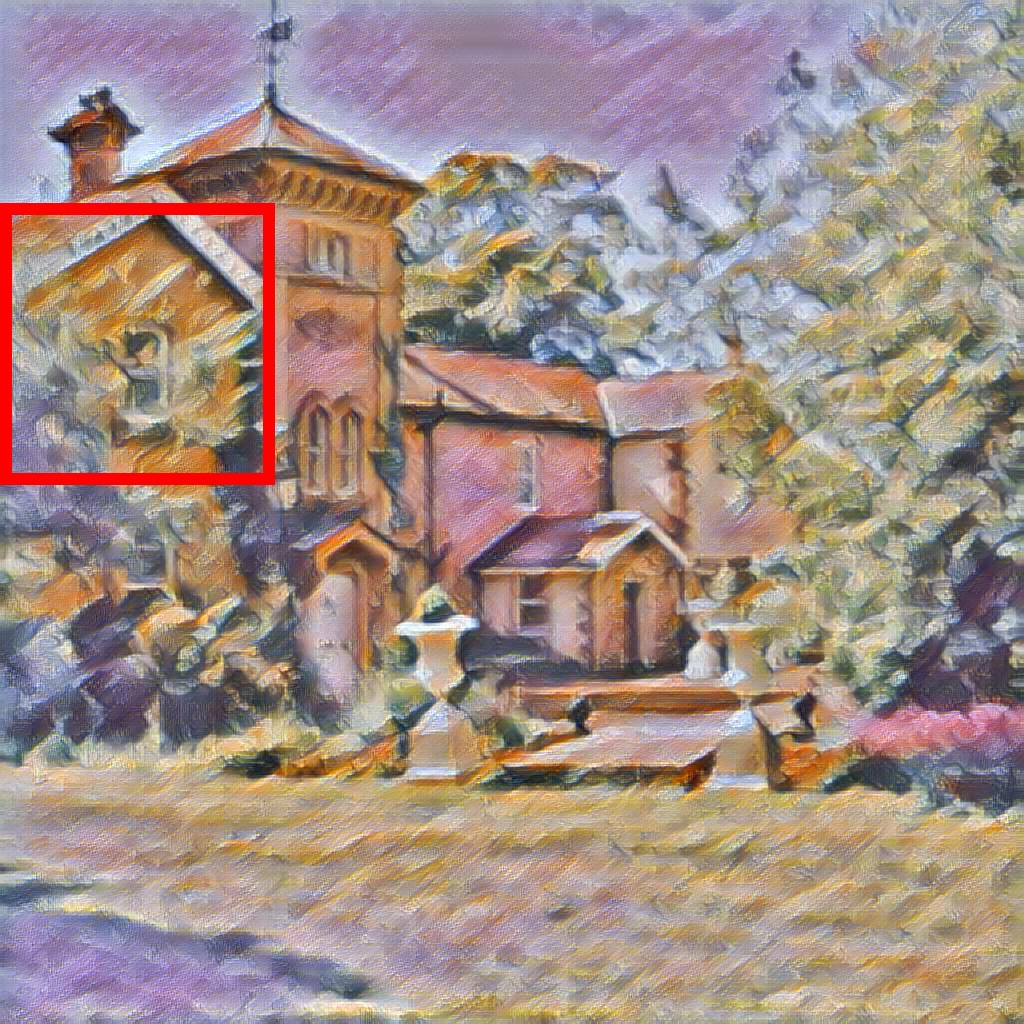} \\
    	\includegraphics[width = 1.1in,height = 1.1in, trim={12 556 764 216},clip]{images/intro_figure_1/day_mt.jpg}
    	\end{tabular}
    } &
    \subfloat[Content]{
    	\begin{tabular}{c}
    	\includegraphics[width = 1.1in]{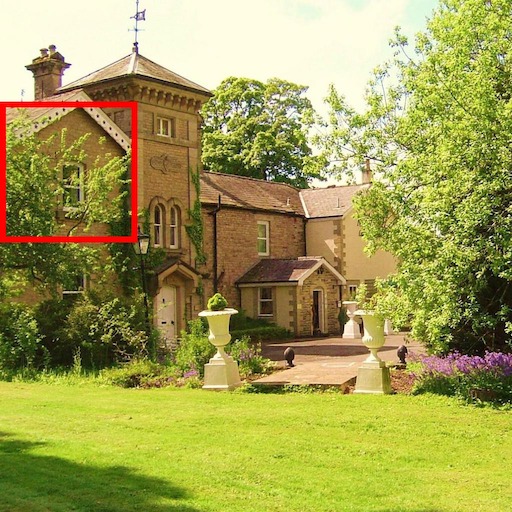} \\
    	\includegraphics[width = 1.1in,height = 1.1in, trim={6 278 382 108},clip]{images/intro_figure_1/house.jpg}
    	\end{tabular}
    }  
    \end{tabular}
    
\end{center}
\vspace*{-2ex}
   \caption{Top row: (a) The style guide  is \textit{At the Close of Day} by Tomas King, and (f) is the content image. (b) is the result of Gatys \textit{et al}'s optimization-based method. (Result size is 512 due to memory limitation of the method) (c), (d) and (e) are results generated by different feed-forward networks (all are of size 1024). 
   Bottom row: the zoom-in display of the regions enclosed in the  red boxes from  the top row. As can be seen, all results are repainted with the color of the style image. 
   However, a closer examination shows that the brush strokes are  not captured  well in (c) and (d). 
   The zoom-in region in (b) is a little blurry. Comparing with the others, our multimodal transfer (e) is capable of simulating more closely the brushwork of the original artwork on high-resolution images.}
 \vspace*{-1ex}
\label{fig:intro}
\end{figure*}

Style transfer, or to repaint an existing photograph with the style of another, is considered a challenging but interesting problem in arts. 
Recently, this task has become an active topic both in academia and industry due to the influential work by Gatys \textit{et al.}~\cite{gatys2016image}, where a pre-trained deep learning network for visual recognition is used to capture both style and content representations, and achieves visually stunning results. 
Unfortunately, the transfer run time is prohibitively long because of the online iterative optimization procedure. 
To resolve this issue, a feed-forward network can be trained offline with the same loss criterion to generate stylized results that are visually close (but still somewhat inferior). 
In this way, only one single inference pass of the feed-forward network is needed at the application time. 
This results in a computational algorithm that is hundreds of times faster~\cite{johnson2016perceptual,ulyanov2016texture}.

Though past work creates visually pleasing results for many different types of artworks, two  important drawbacks stand out: 
(1) The current feed-forward networks ~\cite{johnson2016perceptual,ulyanov2016texture} are trained on a specific resolution of the style image, so deviating from that resolution (bigger or smaller) results in a scale mismatch. 
For example, applying a model trained with a style guide of size 256 on higher-resolution images would generate results whose texture scale is smaller than that of the artistic style,  and 
(2) Current networks often fail to capture small, intricate textures, like brushwork, of many kinds of artworks on high-resolution images. 
While it has been shown that these feed-forward networks function quite well on artworks with abstract, large-scale textures and  easily discernible strokes, \eg, \textit{The Starry Night} by Vincent van Gogh, artistic styles are much more encompassing than what has been demonstrated. 
That is, different artistic styles may be characterized by exquisite, subtle brushes and strokes, and hence, our observations are that results of these style-transfer networks are often not satisfactory for a large variety of artistic styles.   

In this paper, we propose a novel hierarchical deep convolutional neural network architecture for fast style transfer. Our contribution is fourfold: 
(1) We introduce a hierarchical network and design an associated training scheme that is able to learn both coarse, large-scale texture distortion and fine,  exquisite brushwork of an artistic style by utilizing multiple scales of a style image;
(2) Our hierarchical training scheme and end-to-end CNN network architecture allow us to combine multiple models into one network to handle increasingly larger image sizes; 
(3) Instead of taking only RGB color channels into consideration, our network utilizes representations of both color and luminance channels for style transfer;
and  (4) Through experimentation, we show that our hierarchical style transfer network can better capture both coarse and intricate texture patterns. 

Our hierarchical style transfer network is trained with multiple stylization losses at different scales using a mixture of  modalities,  so we distinguish it as \textit{multimodal transfer} from the feed-forward style transfer networks with only one stylization loss \cite{johnson2016perceptual,ulyanov2016texture}, which we call \textit{singular transfer}. 
In Fig.\ \ref{fig:intro} we give an example that compares results from our multimodal transfer network with those from the current state-of-the-art singular transfer networks.
Fig.\ \ref{fig:intro} shows the advantages of multimodal transfer on learning different levels of textures, including style, color, large texture distortion and fine brushwork.
Note specifically that our method can simulate more closely the brushwork of the artwork. In Sec.~\ref{sec:experiments}
we will show that multimodal transfer can also be used to train a combination model to stylize a single image with multiple, distinct artistic styles. 


\begin{figure*}
\vspace*{-1ex}
\begin{center}
\includegraphics[width=6.8in]{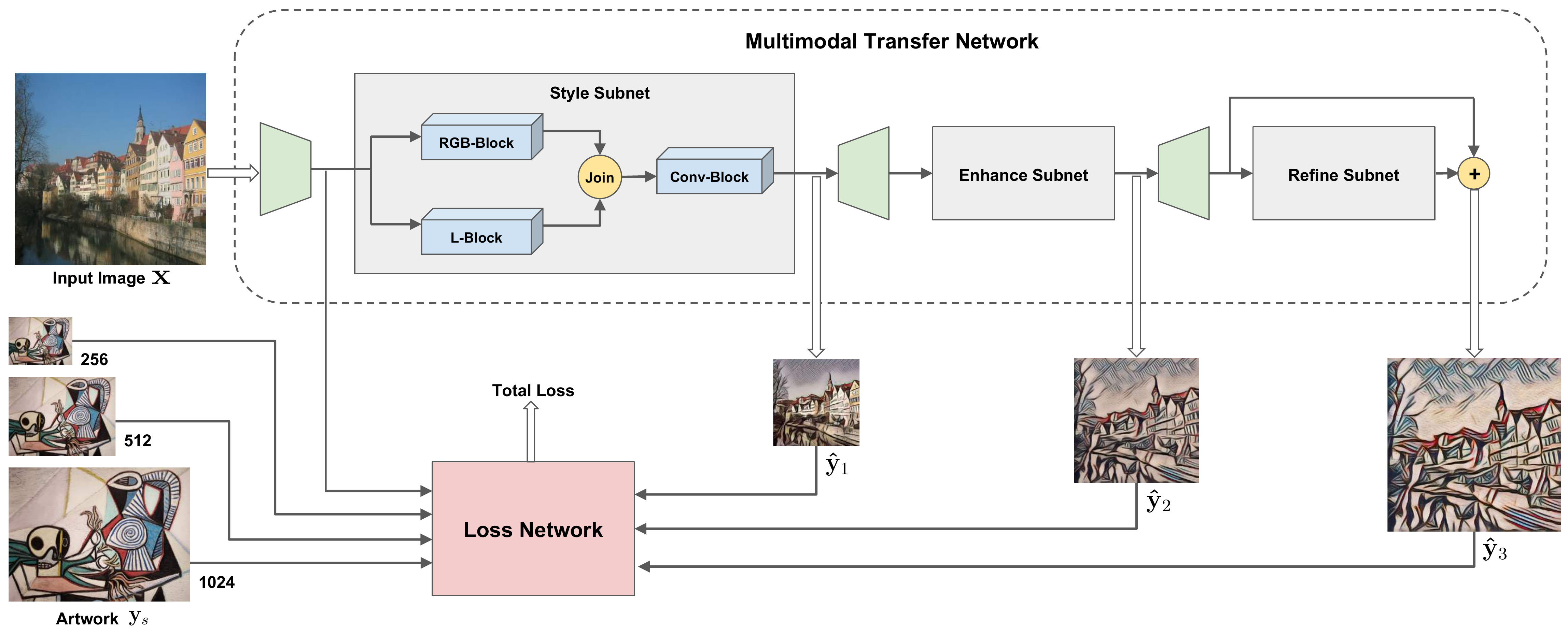}  
\end{center}
\vspace*{-3ex}
   \caption{ Overall Architecture. Please see Sec.~\ref{sec:framework} for explanation.}
\label{fig:overview}
\vspace*{-1ex}
\end{figure*}

\section{Related Work}
\label{sec:review}

\paragraph{Understanding representations of deep neural networks.}
Recently,  seminal work was done on understanding deep neural networks. 
The DeconvNet method of Zeiler and Fergus~\cite{zeiler2014visualizing} learns how certain network outputs are obtained by identifying which image patches are responsible for certain neural activation. 
Yosinski \textit{et al.}~\cite{yosinski2015understanding} aims to understand what computation is performed by deep networks through visualizing the internal neurons.  
Mahendran and Vedaldi~\cite{mahendran2015understanding} inverts the image representations of certain layers to learn what information is preserved by the networks. 
The latter two approaches generate visualization images with an optimization procedure whose objective is for perceptual understanding of network functions. 
Similar optimization procedure is also adopted in other cases~\cite{simonyan2013deep,nguyen2015deep}.

Based on the better understanding of the powerful representations of deep convoluntional  networks~\cite{krizhevsky2012imagenet}, many traditional vision tasks have been addressed with much more improved outcomes.
Optimization-based style transfer is one such example.
Different from previous texture synthesis algorithms that are usually non-parametric methods ~\cite{efros1999texture,wei2000fast,efros2001image,hertzmann2001image,ashikhmin2003fast,kwatra2003graphcut,lee2010directional}, Gatys \textit{et al.} first proposed an optimization method of synthesizing texture images where the objective loss is computed based on the representations of a pre-trained convolutional neural network~\cite{gatys2015texture}. 
This texture loss is then combined with  content loss derived from Mahendran and Vedaldi~\cite{mahendran2015understanding} to perform the style transfer task~\cite{gatys2016image}.

\paragraph{Feed-forward networks for image generation.}
The optimization-based methods for image generation are computationally expensive due to the iterative optimization procedure.
On the contrary, many deep learning methods use the perceptual objective computed from a neural network as the loss function to construct feed-forward neural networks to synthesize images~\cite{dosovitskiy2015learning,goodfellow2014generative,denton2015deep,radford2015unsupervised}. 

Fast style transfer  has achieved great results and is receiving a lot of attention. Johnson \textit{et al.}~\cite{johnson2016perceptual} proposed a feed-forward network for both fast style transfer and super-resolution using the perceptual losses defined in Gatys \textit{et al.}~\cite{gatys2016image}. 
A similar architecture \textit{texture net} is introduced to synthesize textured and stylized images~\cite{ulyanov2016texture}. 
More recently, Ulyanov \textit{et al.}~\cite{ulyanov2016instance} shows that replacing spatial batch normalization~\cite{ioffe2015batch} in the feed-forward network with instance normalization can significantly improve the quality of generated images for fast style transfer. 
Here we present further improvement of such style transfer algorithms to handle progressively larger images using hierarchical networks with mixed modalities. 
Furthermore, it allows the use of multiple, distinct styles for repainting a single input image. 


\section{Multimodal Transfer Network}
\label {sec:MT}

\subsection{Overall Architecture and Learning Schemes}
\label{sec:framework}

Our proposed network, which is shown in Fig.~\ref{fig:overview}, is comprised of two main components: a feed-forward multimodal network and a loss network. 
The feed-forward multimodal network (\textit{MT Network}) is a hierarchical deep residual convolutonal neural network. 
It consists of three subnetworks: \textit{style subnet}, \textit{enhance subnet} and \textit{refine subnet}. These subnets are parameterized by $\Theta_{1}$, $\Theta_{2}$, and $\Theta_{3}$ respectively (these parameters will be made explicit later). 
At a high level, the MT Network takes an image  $\vec{x}$ as the input and is trained to generate multiple output images $\vec{\hat{y}}_k$ of increasing sizes, 
\begin{equation} \label{eq:output}
	\vec{\hat{y}}_k = f(\cup_{i=1}^{k} \Theta_{i}, \vec{x}) \quad .
\end{equation}

These output images are then taken separately as inputs to the loss network  to calculate a stylization loss for each. 
The total loss is a weighted combination of all stylization losses. 
We will show later in Sec.~\ref{sec:loss function} the loss network and the definition of the total loss.  

At test time, in order to produce the same stylization effect and correct texture scale of the artworks when applied to larger images, the MT network stylizes the image hierarchically: The input image is first resized into 256 with a bilinear downsampling layer and stylized by the \textit{style subnet}, capturing the large color and texture traits of the artwork. Next the stylized result, which is the first output $\vec{\hat{y}}_1$, is upsampled into 512 and transferred to the output $\vec{\hat{y}}_2$ by the \textit{enhance subnet}, which enhances the stylization strength. Then it is resized back to 1024. Finally, the \textit{refine subnet} removes the local pixelization artifacts and further refines the result. The high-resolution and most visually appealing result $\vec{\hat{y}}_3$ is obtained after these three-stage processing. Note that while we illustrate the process using a two-level hierarchy, the same concept can be extended recursively to enable stylization of progressively larger images.

\subsection{Loss Functions} \label{sec:loss function}

In this section, we first introduce the single stylization loss funtion and then present a hierarchical stylization loss function that is adopted to train our multimodal transfer network. 

\vspace*{-1ex}
\subsubsection{Single Stylization Loss Function}

\begin{figure}
\vspace*{-1ex}
\begin{center}
\includegraphics[width=3in]{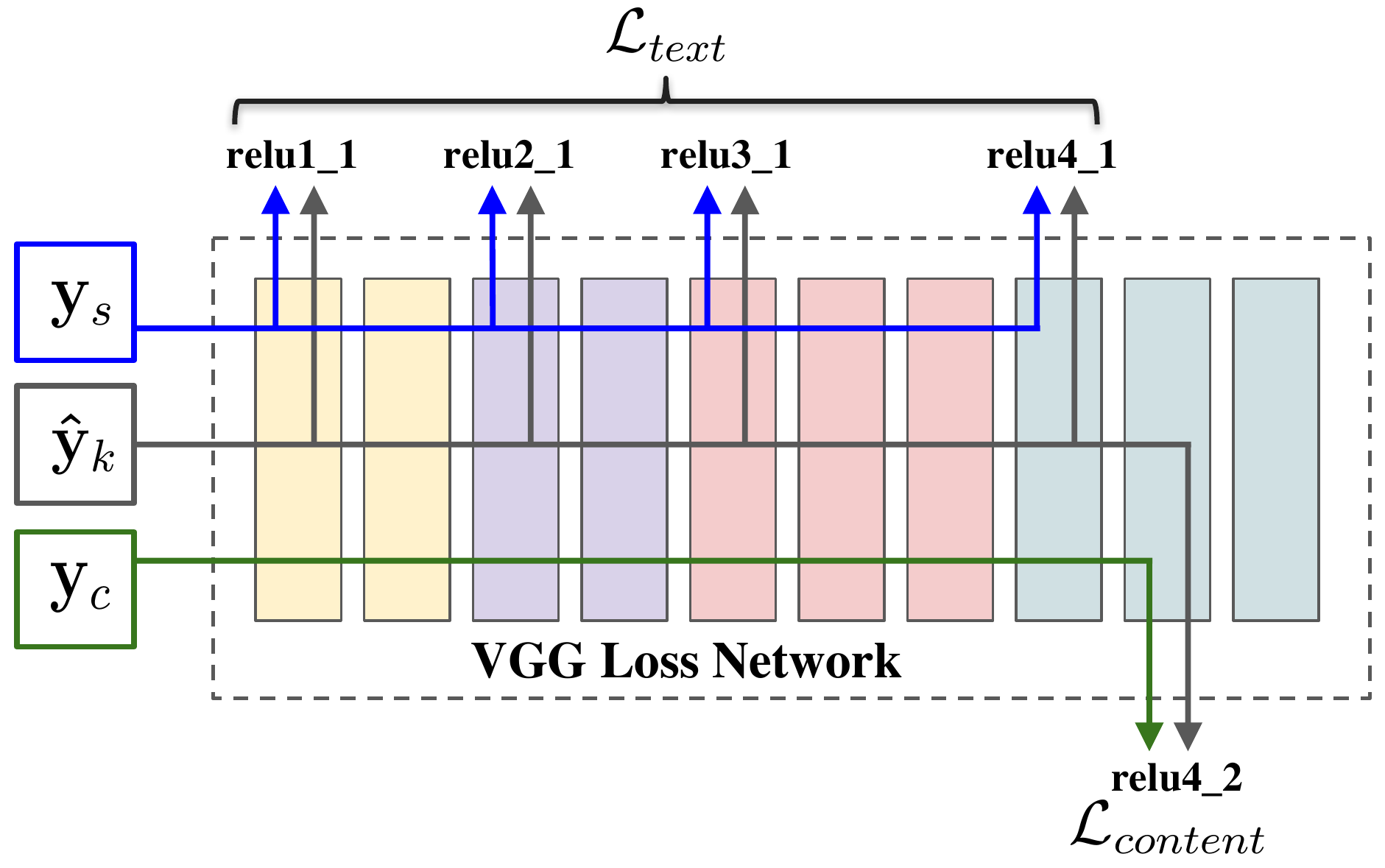}  
\vspace*{-3ex}
\end{center}
   \caption{Loss network. Please see Sec.~\ref{sec:loss function} for explanation.}
\label{fig:loss}
\vspace*{-1ex}
\end{figure}

Similar to the loss definition in previous work for fast style transfer~\cite{johnson2016perceptual,ulyanov2016texture}, the stylization loss is also derived from Gatys \textit{et al.}~\cite{gatys2016image}, where a loss network (a pre-trained VGG-19 network optimized for object recognition~\cite{simonyan2014very}) is used to extract the image representations.  

Two perceptual losses are defined to measure to what extent the generated image $\vec{\hat{y}}_k$ combines the content of the content target $\vec{y}_c$ with the texture and style cues of the style target $\vec{y}_s$ (see Fig. ~\ref{fig:loss}). 

\paragraph {Content Loss}

The content loss function is used to measure the dissimilarity between $\vec{\hat{y}}_k$ and $\vec{y}_c$. 
Let $F_i^l(\vec{x})$ denote the $i$-th feature map in the $l$-th layer of the loss network applied to image $\vec{x}$. The content loss is the squared-error loss between the two feature representations at layer $l$
\begin{equation}
	\mathcal{L}_{content}(\vec{\hat{y}}_k,\vec{y}_c,l)=\sum_{i=1}^{N_l}\|F_i^l(\vec{\hat{y}}_k)-F_i^l(\vec{y}_c)\|_2^2 \quad .
\end{equation}
That is, the content loss directly compares the feature maps computed from the corresponding layers and thus is suitable for characterizing spatial content similarity.

\paragraph{Texture or Style Loss}

Gatys {\em et al.} propose that the correlations between feature maps in each layer of the loss network can be seen as texture representations of an image~\cite{gatys2015texture,gatys2016image}. Those correlations are given by the Gram matrix, whose elements are pairwise scalar products between those feature maps:
\begin{equation}
	G_{ij}^l(\vec{x})=\langle F_i^l(\vec{x}),F_j^l(\vec{x}) \rangle \quad .
\end{equation}
A set of Gram matrices $G^l,l \in L$ is used as the texture representations, which  discard the spatial information but retain the statistic profiles of color and intensity distribution  of an input image. So the texture loss function is defined as
\begin{equation}
	\mathcal{L}_{text}(\vec{\hat{y}}_k,\vec{y}_s) = \sum_ {l \in L} \| G^l(\vec{\hat{y}}_k) - G^l(\vec{y}_s) \|_2^2 \quad .
\end{equation}

Finally,  the stylization loss for each output $\vec{\hat{y}}_k$ from the MT network is defined as a weighted sum of the content loss and the texture loss
\begin{equation}
	\mathcal{L}_{S}(\vec{\hat{y}}_k,\vec{y}_c,\vec{y}_s) = \alpha \mathcal{L}_{content}(\vec{\hat{y}}_k,\vec{y}_c) + \beta \mathcal{L}_{text}(\vec{\hat{y}}_k,\vec{y}_s) ,
\end{equation}
where $\alpha$ and $\beta$ are the weights of the content loss and texture loss, respectively. 

\subsubsection{Hierarchical Stylization Loss Function} \label{sec:Hierarchical Stylization Loss Function}

The multimodal transfer network can generate $K$ output results of $K$ increasing sizes ($K=3$ in the network shown in Fig.~\ref{fig:overview}). Then a stylization loss is computed for each output result $\vec{\hat{y}}_k$
\begin{equation}
	\mathcal{L}_{S}^k(\vec{\hat{y}}_k,\vec{y}_c^k,\vec{y}_s^k) = \alpha \mathcal{L}_{content}(\vec{\hat{y}}_k,\vec{y}_c^k) + \beta \mathcal{L}_{text}(\vec{\hat{y}}_k,\vec{y}_s^k)
\end{equation}
where $\vec{y}_c^k$ and $\vec{y}_s^k$ are the corresponding content target and style target, which are the input to the subnet that outputs $\vec{\hat{y}}_k$, and are the scaled versions of the artwork $\vec{y}_s$. 
By training the subnets with different style scales, we control the types of artistic features that are learned for different subnets. 
Again, we want to emphasize that the concept can be easily extended for more layers. 

Since such stylization losses are computed based on the  outputs of different layers of the whole network, a total loss (\eg, a weighted combination of all stylization losses) cannot be used here to directly propagate and update the weights backward. 
Thus, a parallel criterion is adopted so that different stylization losses are used to back-propagate the weights for different ranges of layers. 
We define the hierarchical stylization loss function $\mathcal{L}_{H}$, which is a weighted sum of such stylization losses, as    
\begin{equation}
	\mathcal{L}_{H} = \sum_{k=1}^K \lambda_k \mathcal{L}_S^k(\vec{\hat{y}}_k, \vec{y}_c^k, \vec{y}_s^k) \quad ,
\end{equation}
where $\lambda_k$ is the weight of stylization loss $\mathcal{L}_S^k$. 

Therefore, during the end-to-end learning on natural images $\vec{x} \sim \mathcal{X}$, each subnet denoted by $\Theta_k$ is trained to minimize the parallel weighted stylization losses that are computed from the latter outputs $\vec{\hat{y}}_i$ ($i \geq k$) (latter means it comes later in the feed-forward direction) as in
\begin{multline} \label{eq:min}
	\Theta_k = \\
    \argmin_{\Theta_k} E_{\vec{x} \sim \mathcal{X}} \Bigg[ \sum_{i \geq k}^K  \lambda_i 		\mathcal{L}_S^k(f(\cup_{j=1}^{k} \Theta_{j}, \vec{x}), \vec{y}_c^i, \vec{y}_s^i) \Bigg]
\end{multline}

In practice, suppose the general back-propagation function is denoted by $f^{-1}$, then for every iteration, the weight updates (gradients) of the subnet $\Theta_k$ can be written as
\begin{equation}
	\Delta \Theta_k = \bigg\{ 
	\begin{tabular}{l l}
		$f^{-1}(\lambda_k \mathcal{L}_{S}^k)$ & $k=K$ 	\\
        $f^{-1}(\lambda_k \mathcal{L}_{S}^k, \Delta \Theta_{k+1})$ & $1 \leq k < K$ ~ ,
	\end{tabular}
\end{equation}
so the weights of the current subnet $\Theta_k$ are influenced by both the stylization loss at the current level $\mathcal{L}_{S}^k$ and the gradients of the latter subnets.

From Eq.~\eqref{eq:min}, we can see that even though all those subnets are designed for different purposes, they are not totally independent. 
Former subnets also contribute to minimize  losses of the latter. 
Thus, shallower CNN structure can be used for latter subnets, which saves both computing memory and running time.

\subsection{Network Architecture}
\label{sec:architecture}

One of the key drawbacks of singular transfer networks (\eg \cite{johnson2016perceptual,ulyanov2016texture}) is that the scale at which the singular transfer network is trained limits the range of style details that are captured. 
Since it is trained with one particular scale of the style image, during training we need to choose if it learns the coarse texture or the fine brushwork. 
That is, it learns one at the expense of the other. 

To remedy this problem,  we design the hierarchical architecture where different subnets are trained with different scales of the style image to learn different levels of artistic texture cues. 
This design enables a test image to be transferred using different levels of the style in increasing resolutions. 
Furthermore, because all these subnets are combined into one network and trained hierarchically, the latter subnets are also able to enhance and refine the results from previous ones, making ours a collaborative scheme for improved efficiency and robustness.

We have experimented with several architectures that have varying levels of hierarchy and different internal structures. 
Here we introduce the general architecture of the network shown in Fig.~\ref{fig:overview}, which has the best stylization quality from our experience.

As stated before, the multimodal transfer network consists of three learnable subnetworks, style subnet, enhance subnet and refine subnet, each following a fixed bilinear upsampling/downsampling layer. 
Note that the upsampling layer between enhance subnet and refine subnet is only inserted at test time, so during training the input to refine subnet is still of size 512, which hugely reduces the required memory and speeds up the training process. 
The salient features of these networks are explained below. 


\begin{figure}
\vspace*{-1ex}
\begin{center}
	\setlength{\tabcolsep}{6pt}
	\begin{tabular}{l l}
	\subfloat[Content]{\includegraphics[width = 1in]{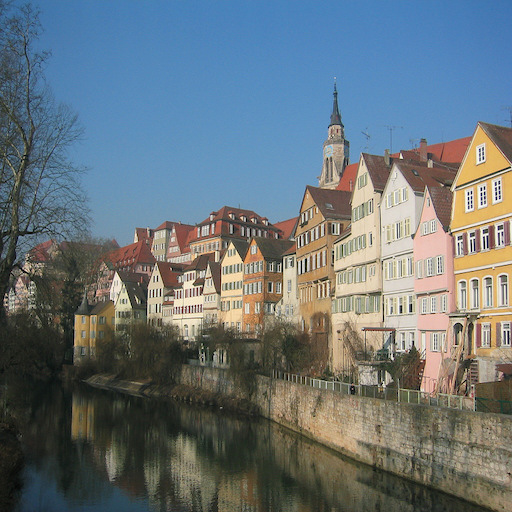}} &
	\subfloat[Style]{\includegraphics[height = 1in]{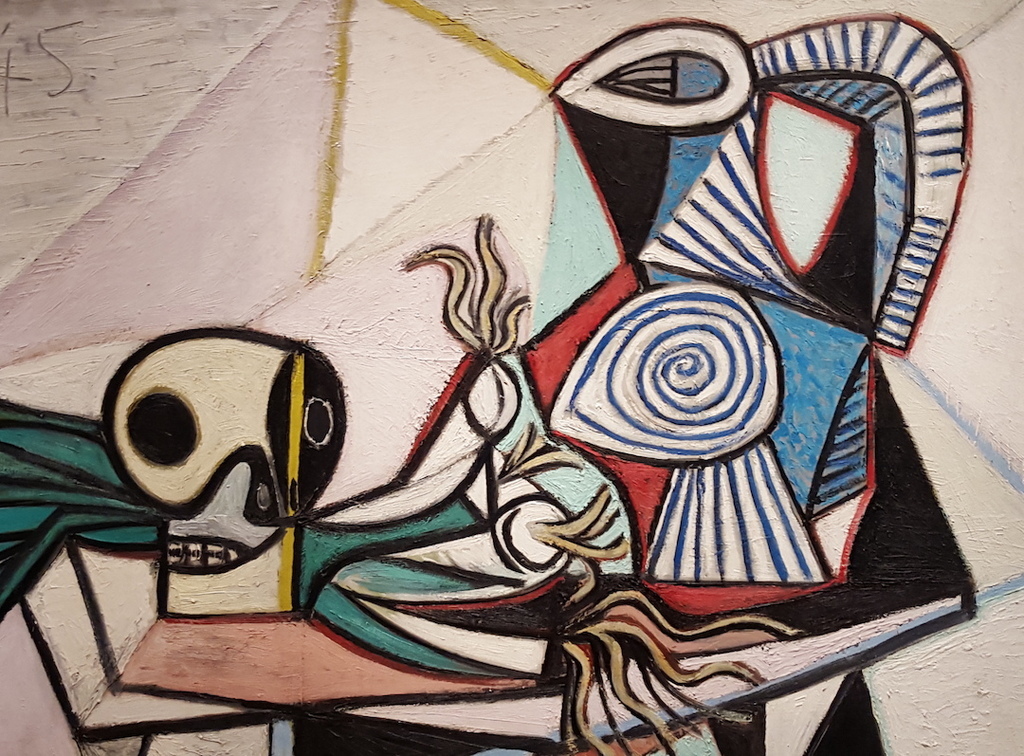}}   
    \end{tabular} 
    \\[-2ex]
    \setlength{\tabcolsep}{2pt}
	\begin{tabular}{c c c}
    \subfloat[]{\includegraphics[width = 1in]{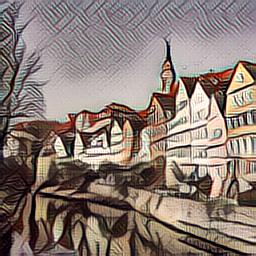}} &
	\subfloat[]{\includegraphics[width = 1in]{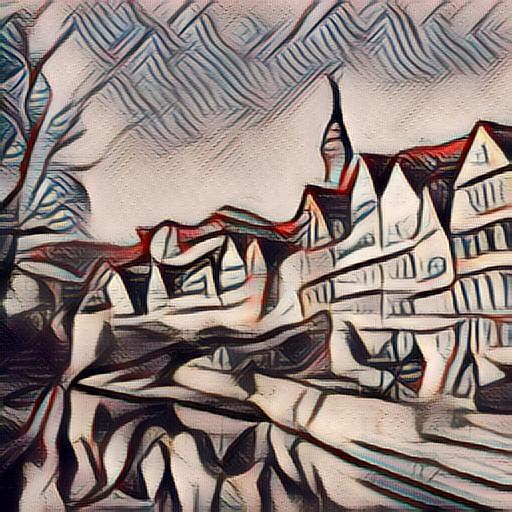}} &
	\subfloat[]{\includegraphics[width = 1in]{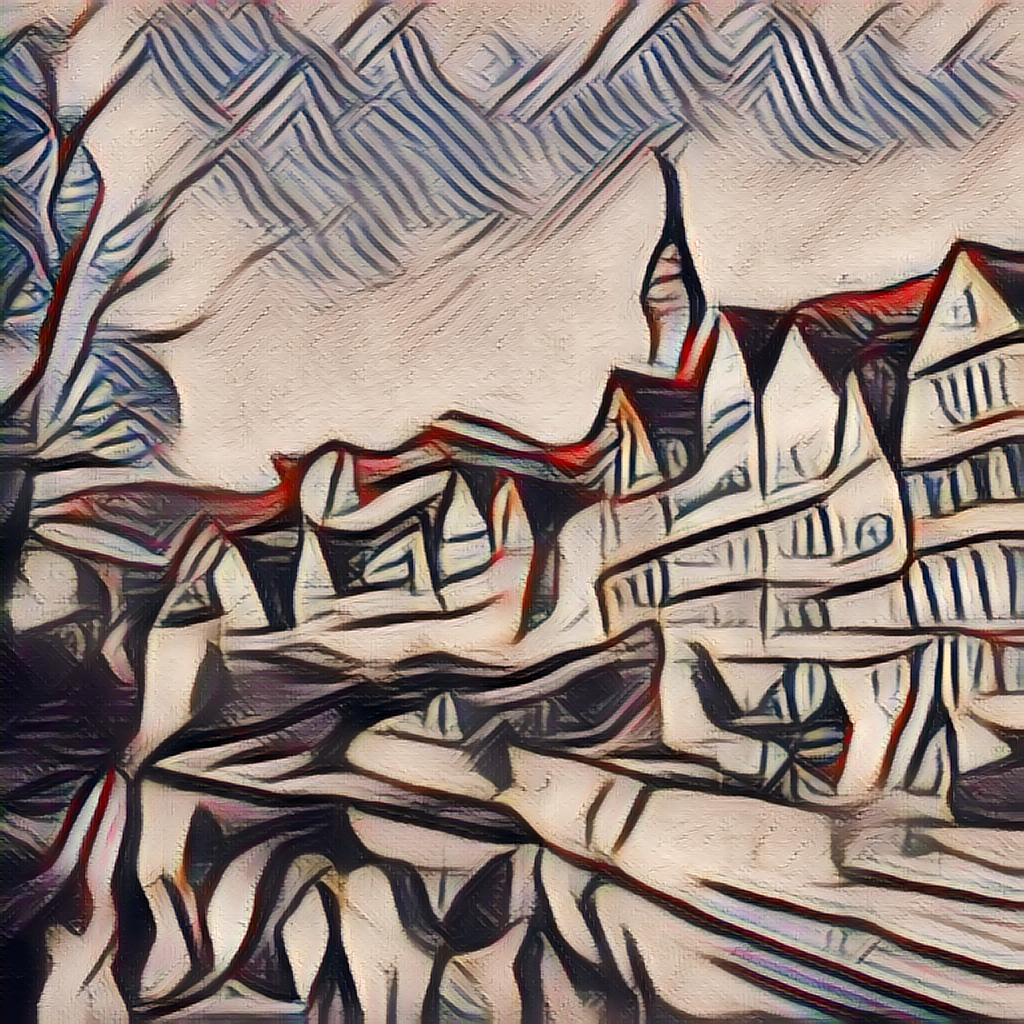}} 
    \\ [-2ex]
    \subfloat[]{\includegraphics[width = 1in]{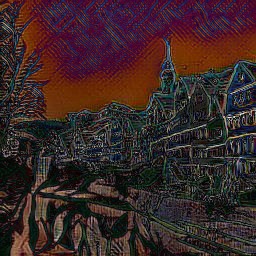}} & 
    \subfloat[]{\includegraphics[width = 1in]{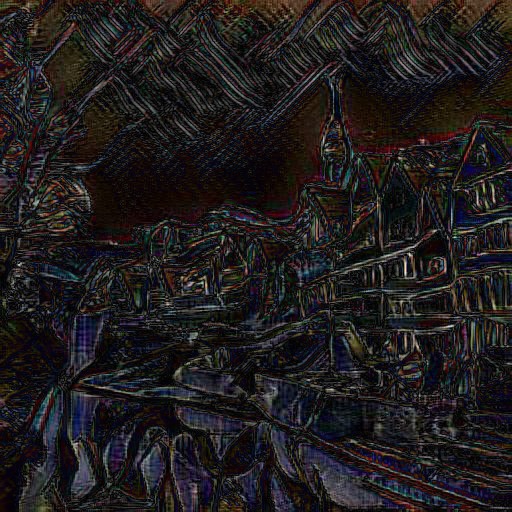}} & 
    \subfloat[]{\includegraphics[width = 1in]{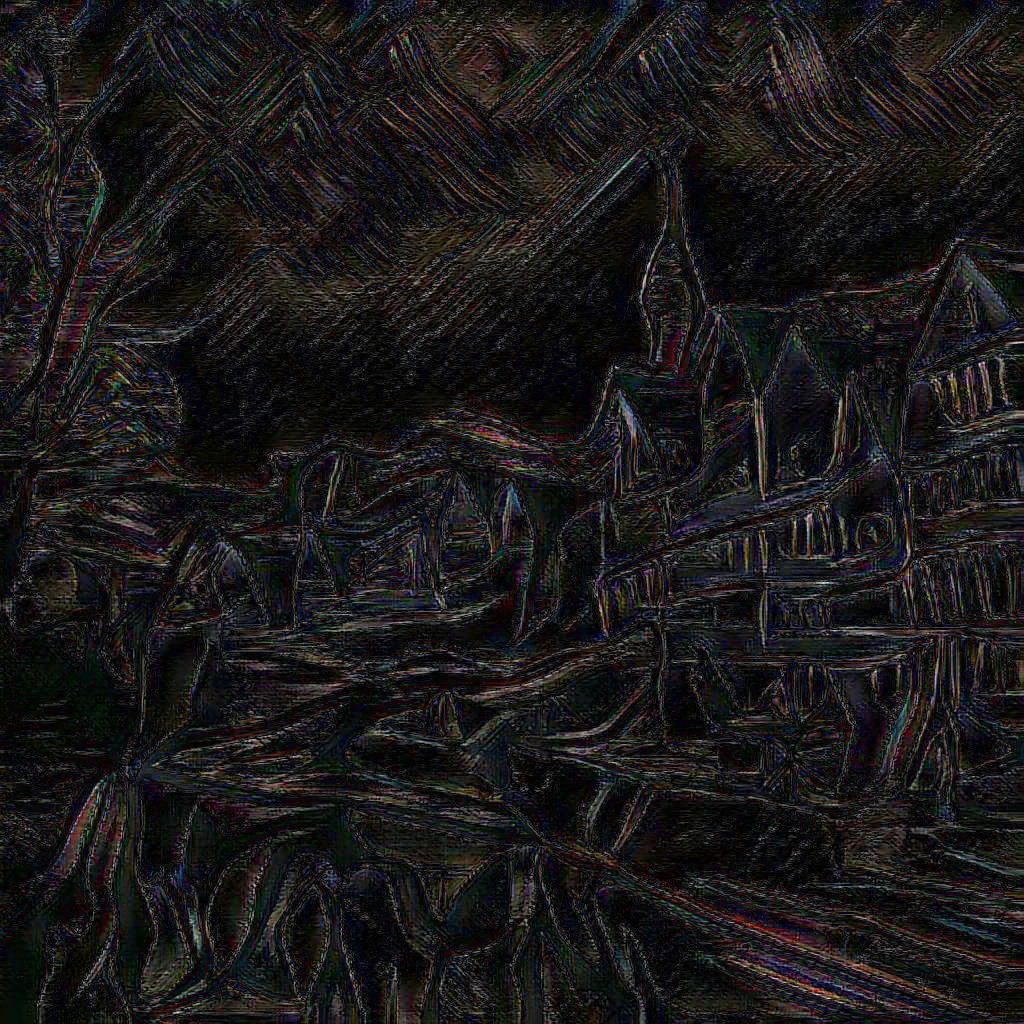}}
    \end{tabular}  
    \vspace*{-3ex}
\end{center}
   \caption{ (a) is the content image and (b) is the style image. (c)(d)(e) show the outputs of the three subnets, $\vec{\hat{y}}_1$, $\vec{\hat{y}}_2$ and $\vec{\hat{y}}_3$, whose sizes are 256, 512 and 1024 respectively. The third row depicts the absolute difference between: (f) the content image and the output image $\vec{\hat{y}}_1$, (g) the output images $\vec{\hat{y}}_1$ and $\vec{\hat{y}}_2$, and (h) the output images $\vec{\hat{y}}_2$ and $\vec{\hat{y}}_3$.}
\label{fig:diff}
\vspace*{-1ex}
\end{figure}

\begin{figure*}
\vspace*{-1ex}
\captionsetup[subfigure]{labelformat=empty}
\begin{center}
	\setlength{\tabcolsep}{0pt}
    \renewcommand{\arraystretch}{0.5}
	\begin{tabular}{c c c c c}
      \includegraphics[width = 1.3in,height=1.3in]{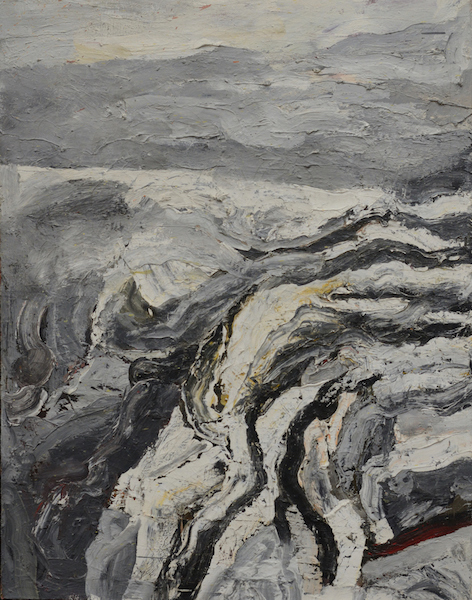} 
      &
      \includegraphics[width = 1.3in]{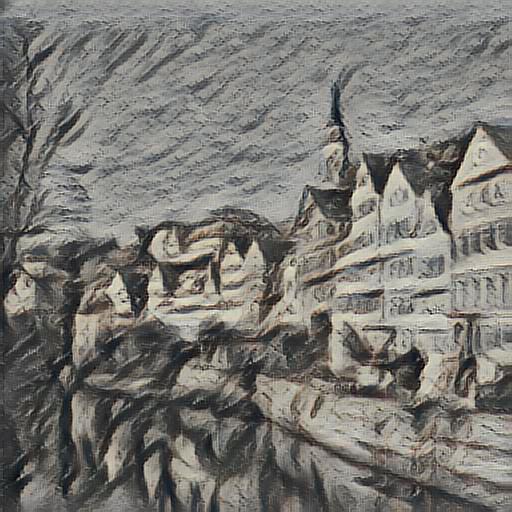} 
      & 
      \includegraphics[width = 1.3in]{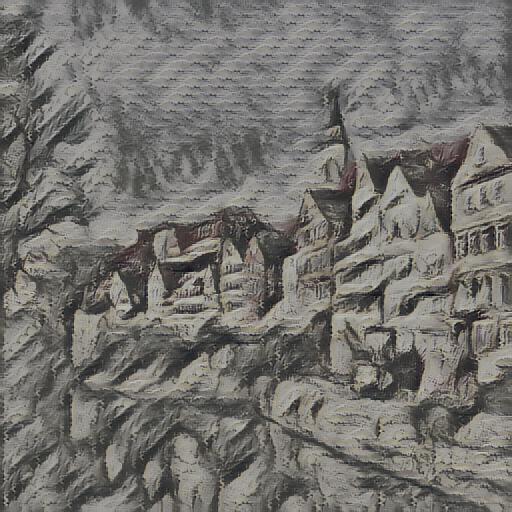} 
   	  &
      \includegraphics[width = 1.3in]{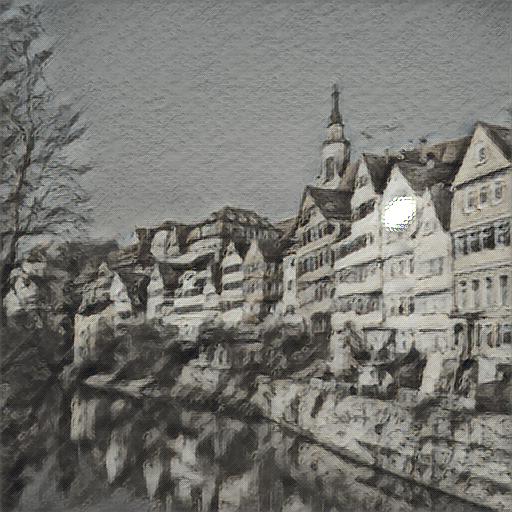} 
      &
      \includegraphics[width = 1.3in]{images/contents/demo_1024.jpg} 
    \\ [-1.5ex]
    \subfloat[(a) Style]{  
      \begin{tabular}{c}
      \includegraphics[width = 1.3in]{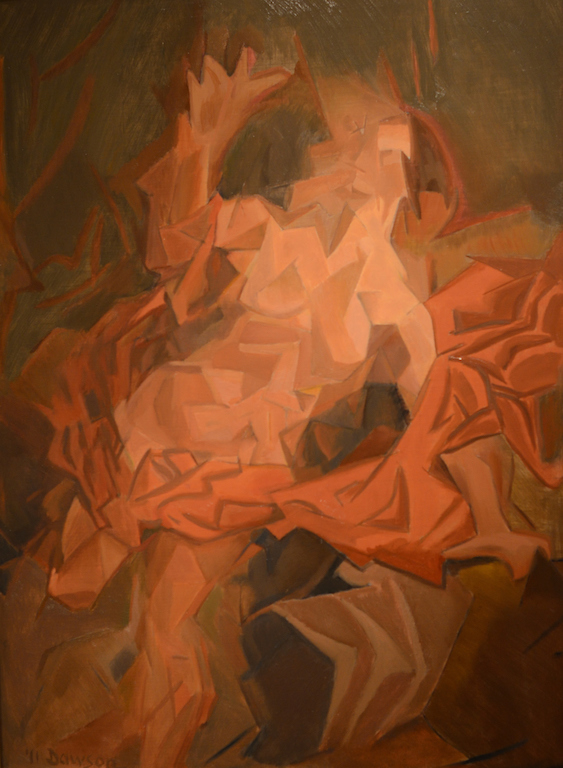} 
      \end{tabular} 
      }&
      \subfloat[ (b) style subnet (ours)]{
      \begin{tabular}{c}
      \includegraphics[width = 1.3in]{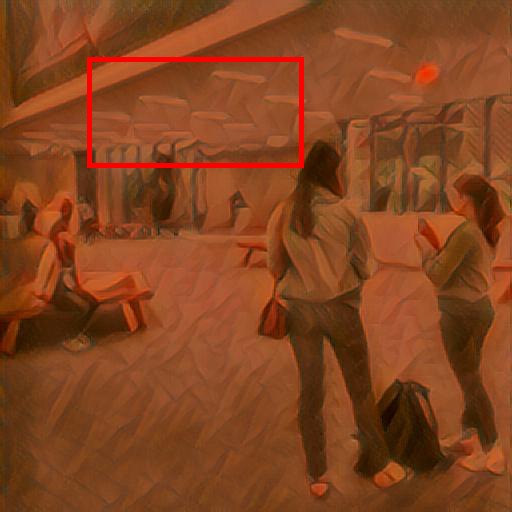} \\
      \includegraphics[width = 1.3in,trim={92 350 214 64},clip]{images/singular/dawson_style_subnet.jpg}
      \end{tabular}
      }& 
      \subfloat[(c) Ulyanov \textit{et al.}]{
      \begin{tabular}{c}
      \includegraphics[width = 1.3in]{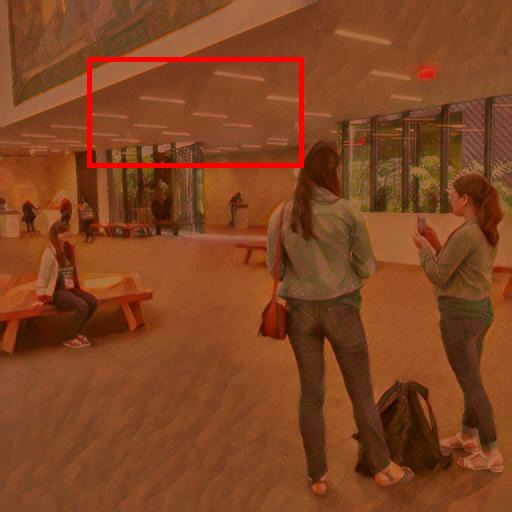} \\
      \includegraphics[width = 1.3in,trim={92 350 214 64},clip]{images/singular/dawson_pyramid.jpg}
      \end{tabular}
   	  }&
      \subfloat[(d) Johnson \textit{et al.}]{
      \begin{tabular}{c}
      \includegraphics[width = 1.3in]{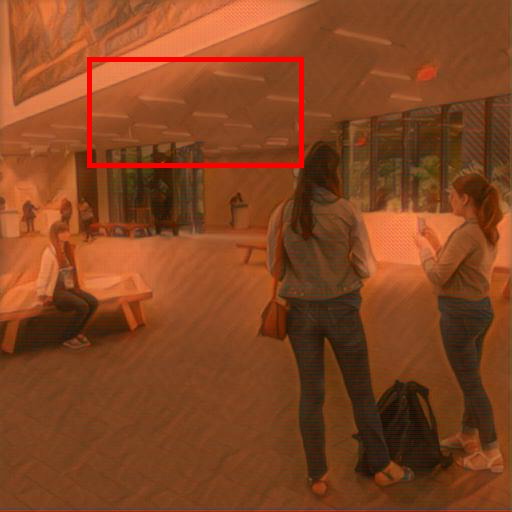} \\
      \includegraphics[width = 1.3in,trim={92 350 214 64},clip]{images/singular/dawson_johnson.jpg}
      \end{tabular}
      }&
      \subfloat[(e) Content]{
      \begin{tabular}{c}
      \includegraphics[width = 1.3in]{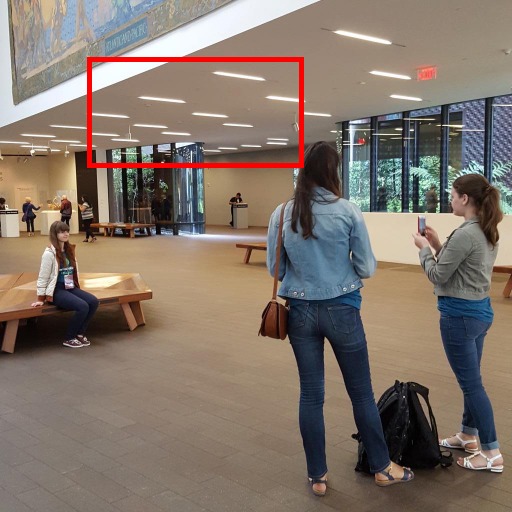} \\
      \includegraphics[width = 1.3in,trim={94 351 214 64},clip]{images/singular/deyoung.jpg}
      \end{tabular}
      }
    \end{tabular}
    \vspace*{-2ex}
\end{center}
   \caption{ Comparison between our style subnet (b) and other singular transfer networks (c) and (d). They were tested on two styles here, \textit{Mountain No. 2} by Jay DeFeo (top) and \textit{Venus} by Manierre Dawson (bottom). All results were of size 512. Notice that for the second style, we also zoomed in on the region in the red box to better compare the fine texture.}
\label{fig:singular}
\vspace*{-1ex}
\end{figure*}

\begin{figure*}
\vspace*{-1ex}
\captionsetup[subfigure]{labelformat=empty}
\begin{center}
	\setlength{\tabcolsep}{2pt}
    \renewcommand{\arraystretch}{0}
	\begin{tabular}{c c c c}
	\subfloat{\includegraphics[width = 1.7in,height = 1.7in]{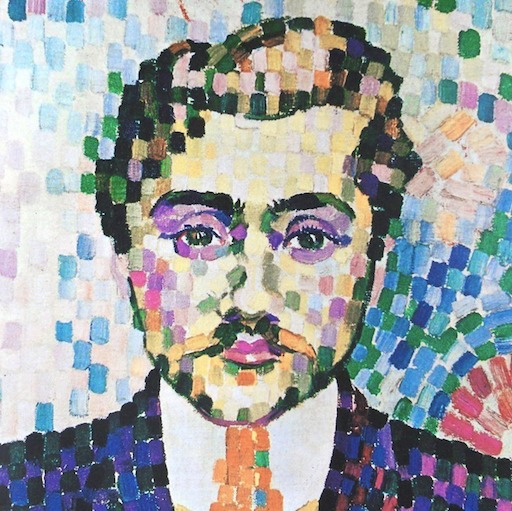}} & 
    \subfloat{\includegraphics[width = 1.7in]{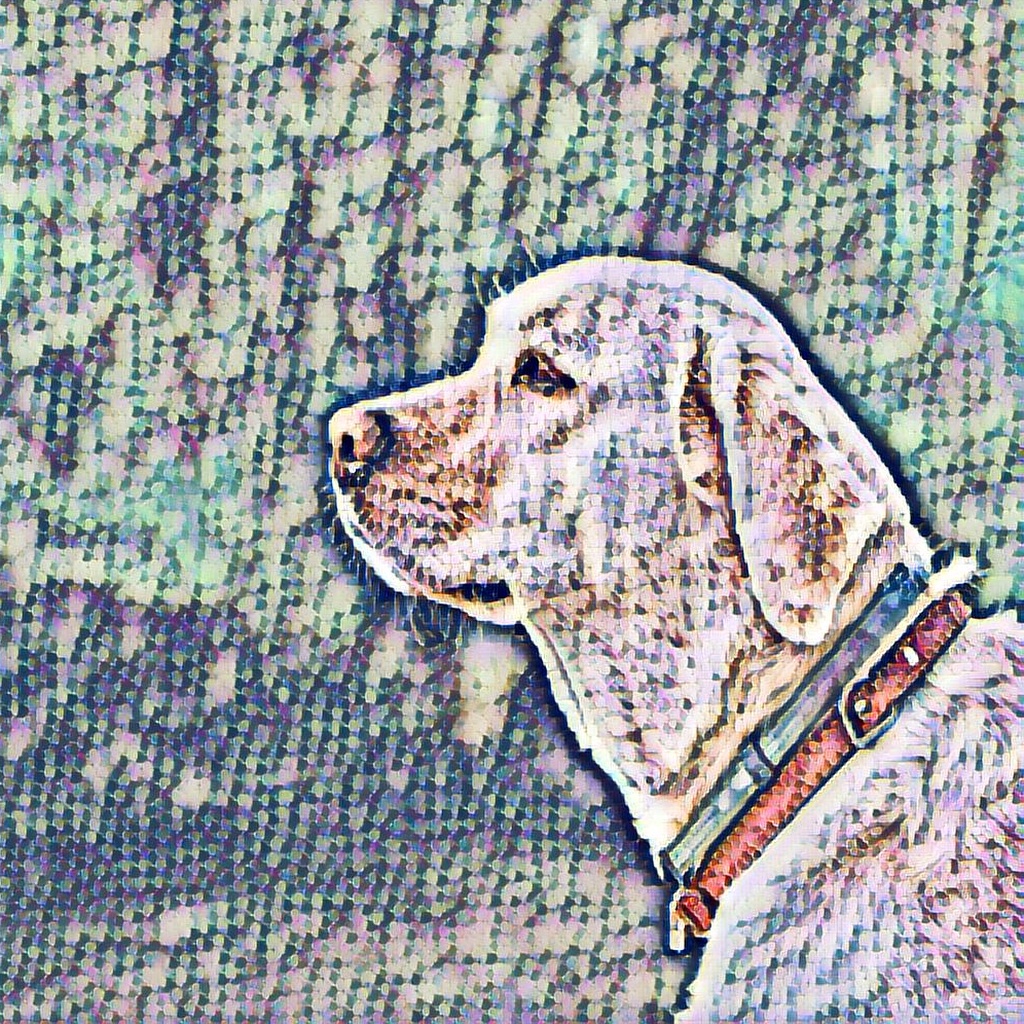}} & 
    \subfloat{\includegraphics[width = 1.7in]{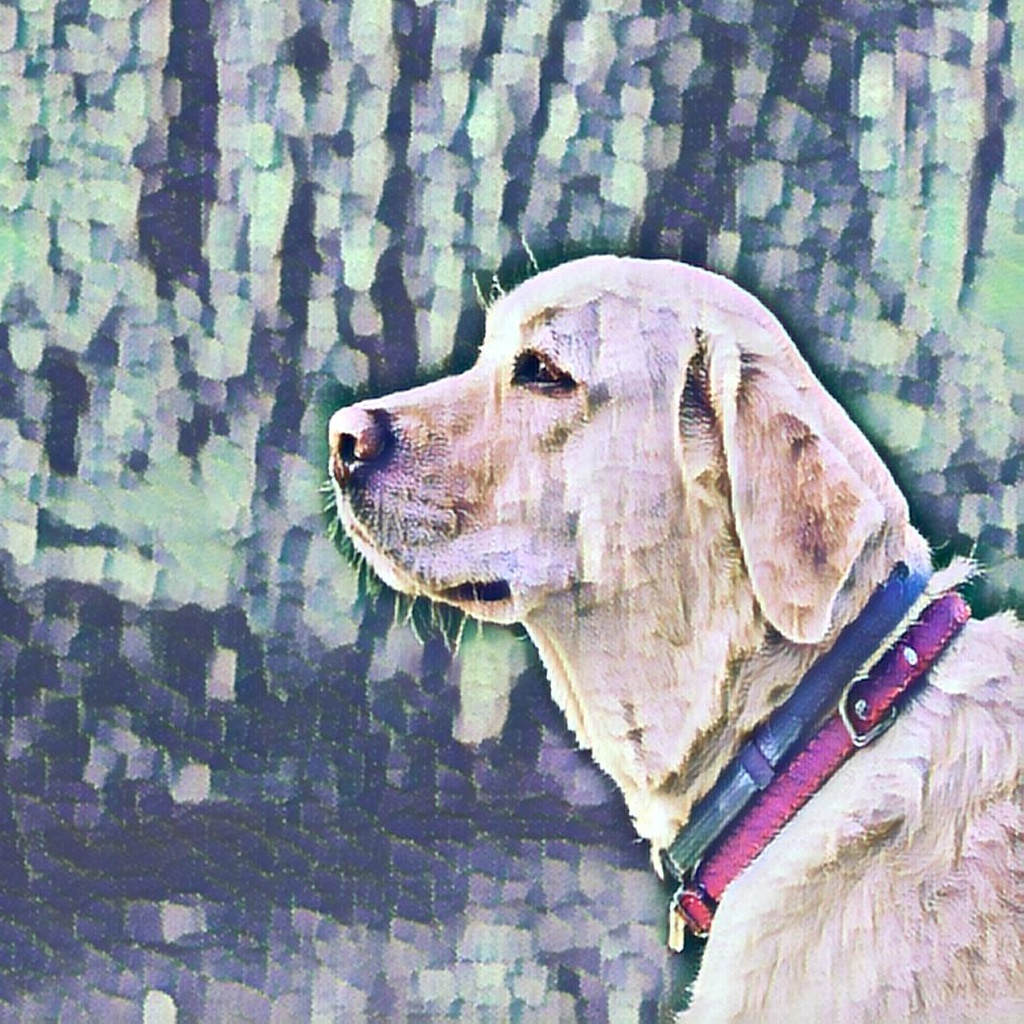}} &
    \subfloat{\includegraphics[width = 1.7in]{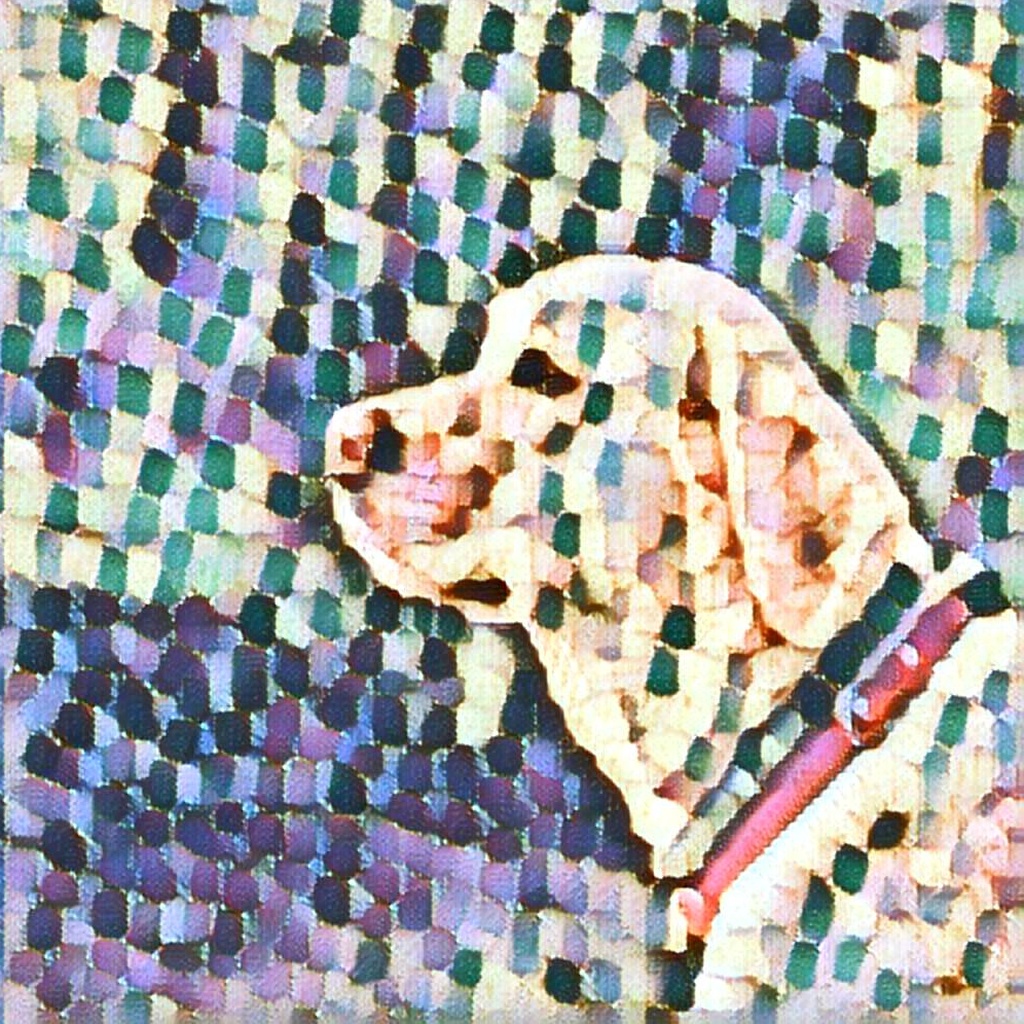}}
    \\ [-1ex]
    \subfloat{\includegraphics[width = 1.7in,height = 1.7in]{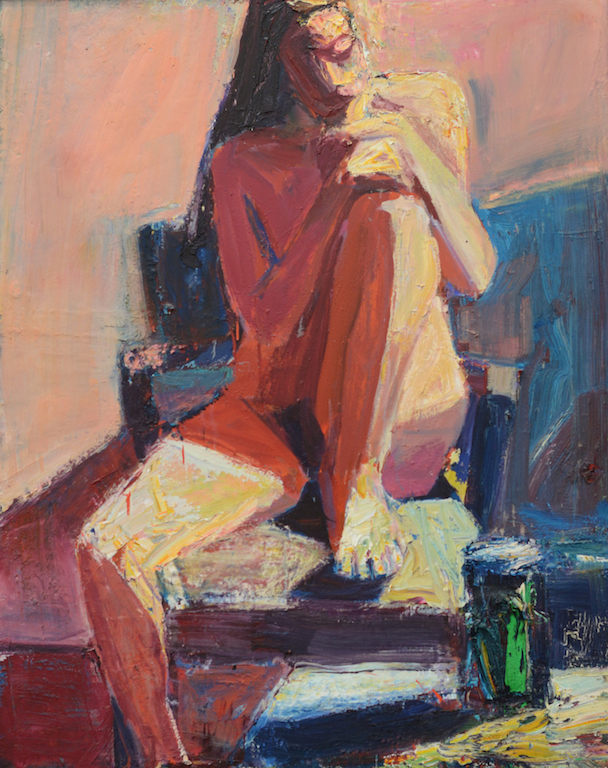}} & 
    \subfloat{\includegraphics[width = 1.7in]{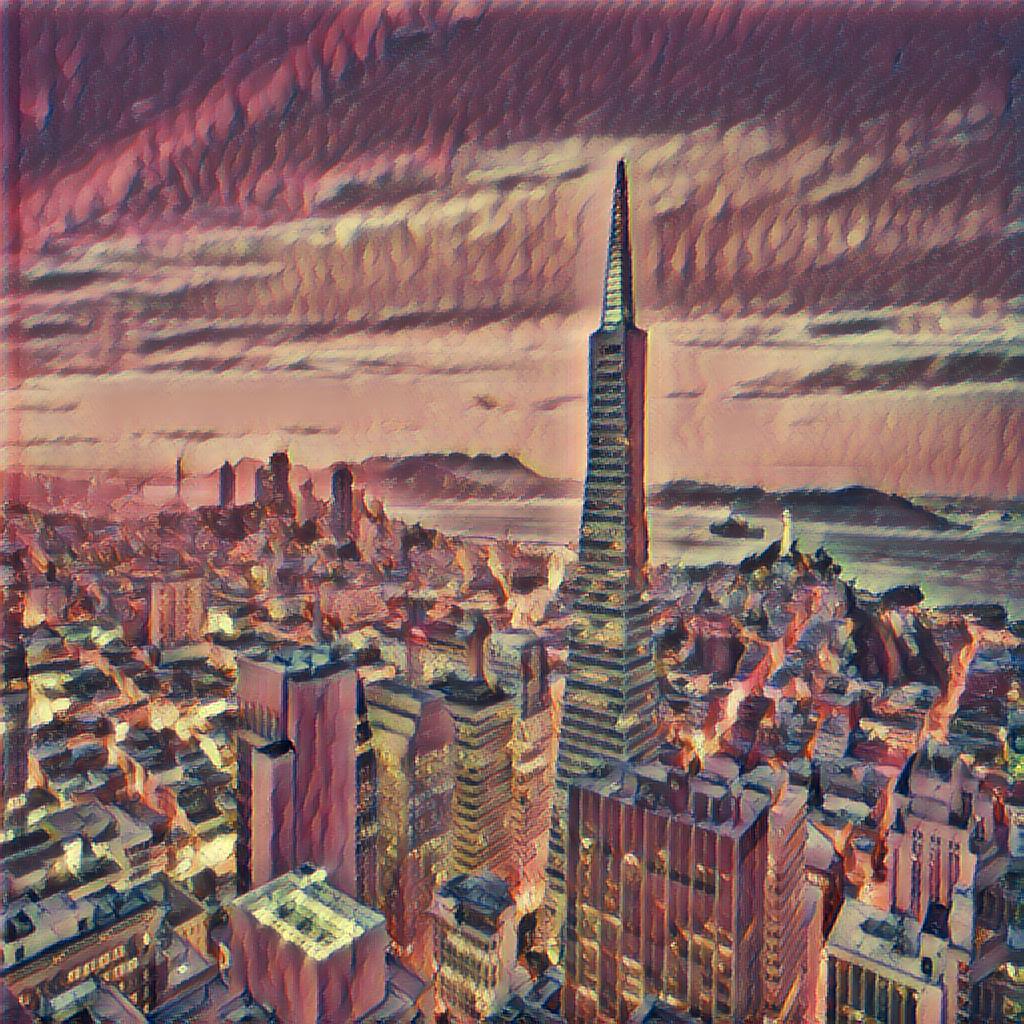}} & 
    \subfloat{\includegraphics[width = 1.7in]{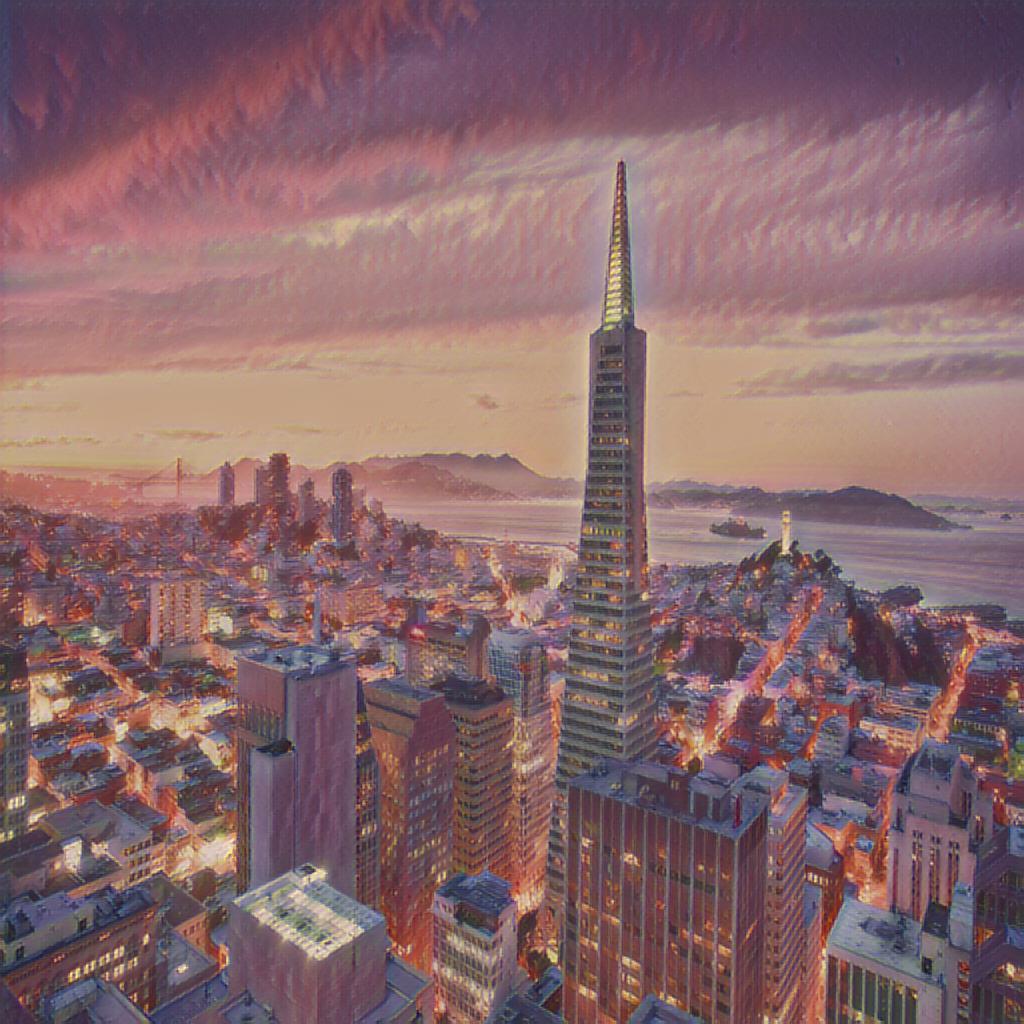}} &
    \subfloat{\includegraphics[width = 1.7in]{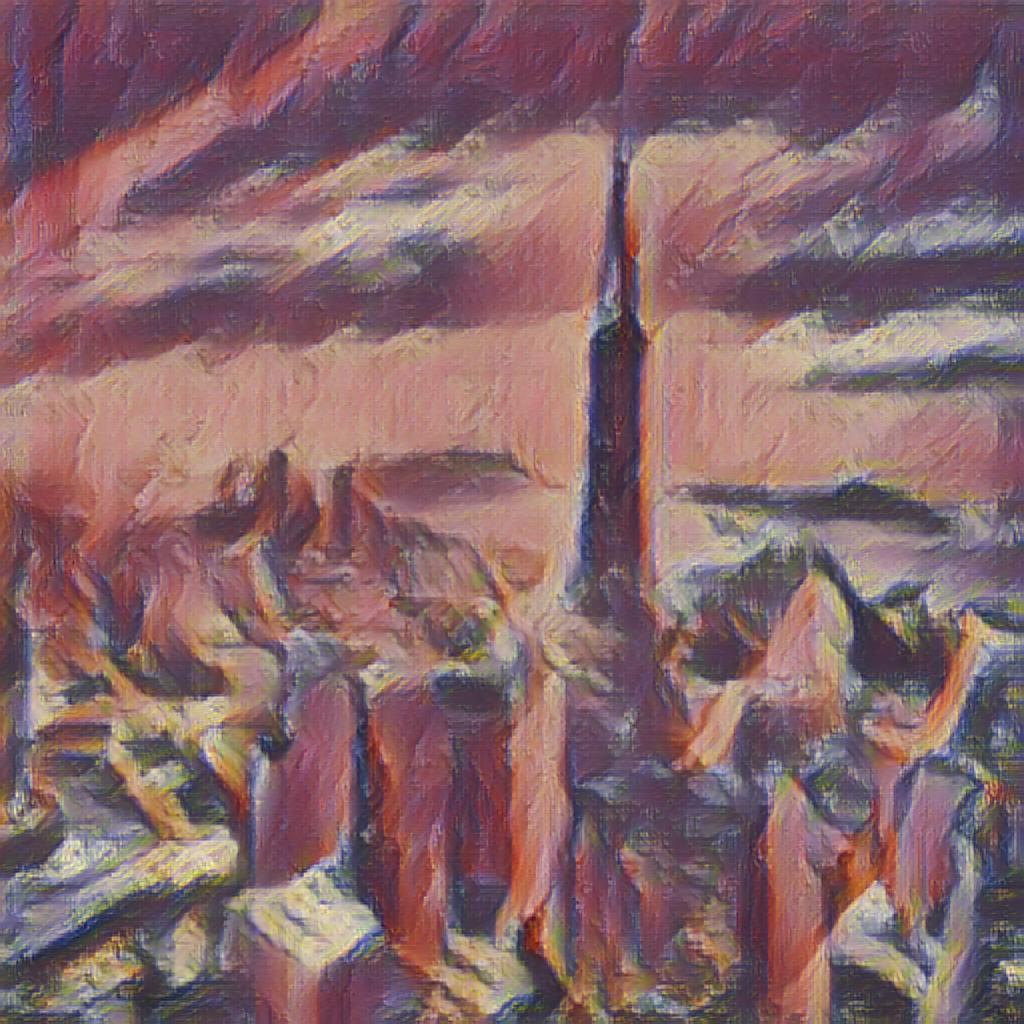}}
    \\ [-1ex]
    \subfloat{\includegraphics[width = 1.7in, height = 1.7in]{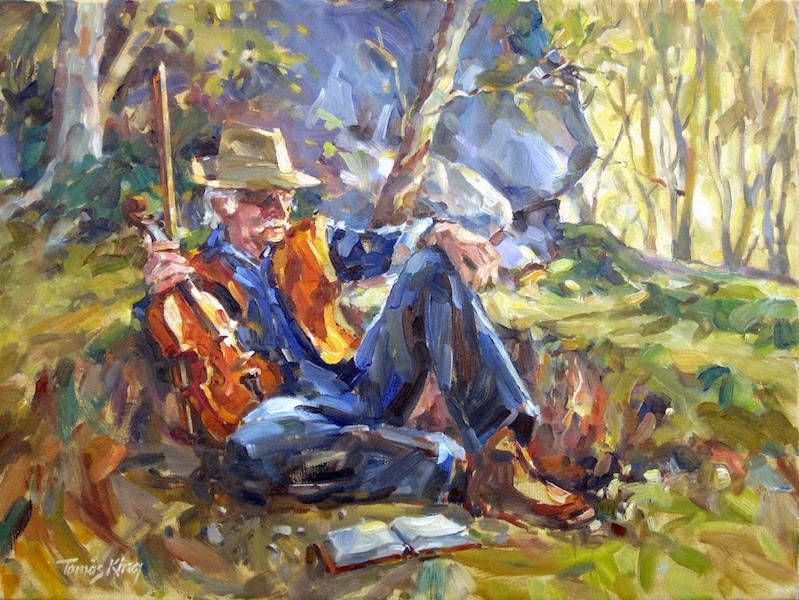}} & 
    \subfloat{\includegraphics[width = 1.7in]{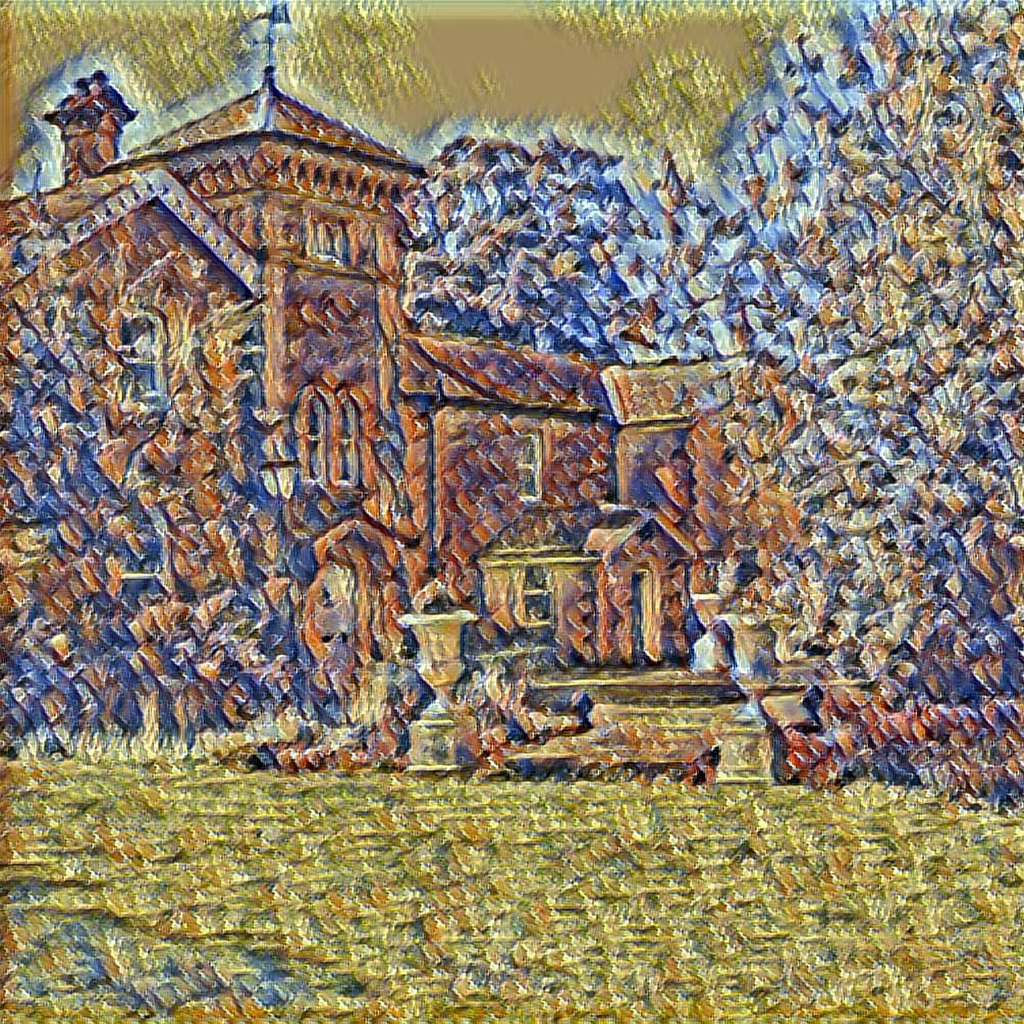}} & 
    \subfloat{\includegraphics[width = 1.7in]{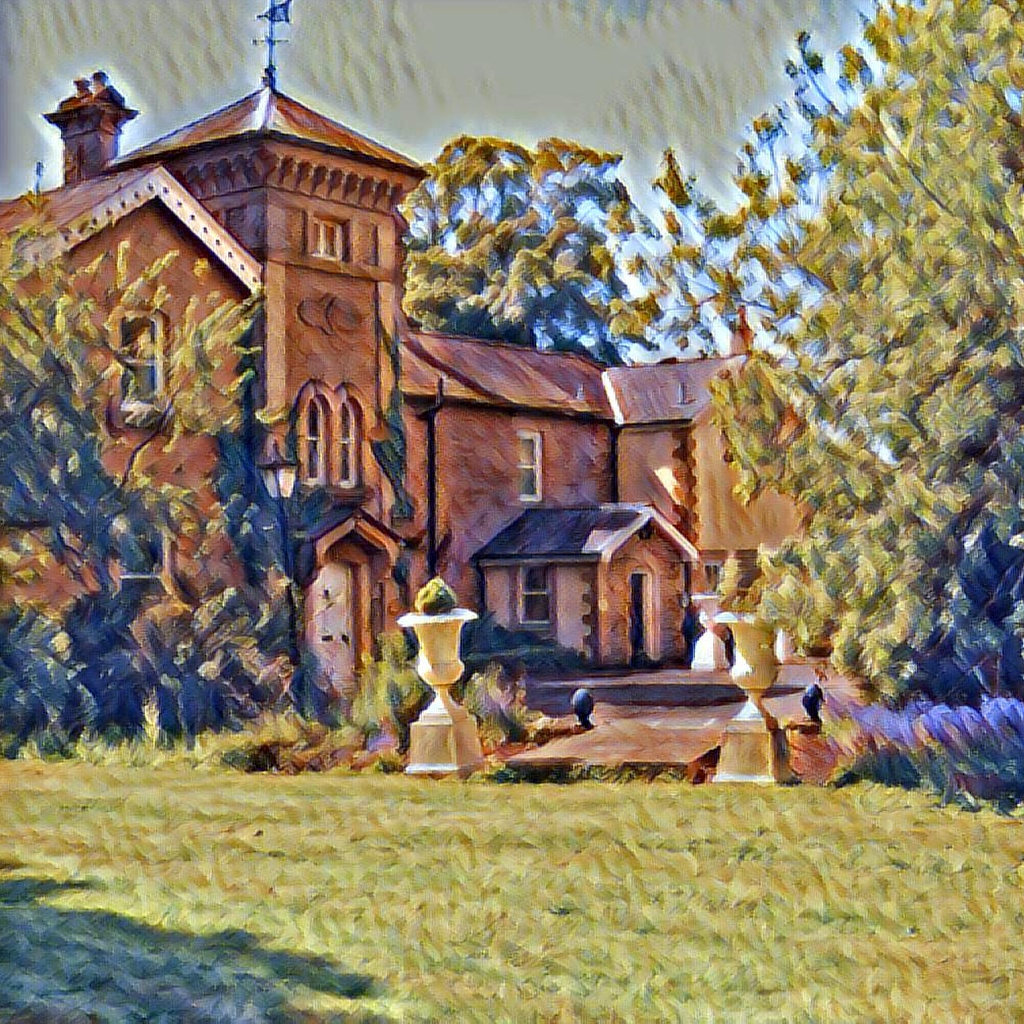}} &
    \subfloat{\includegraphics[width = 1.7in]{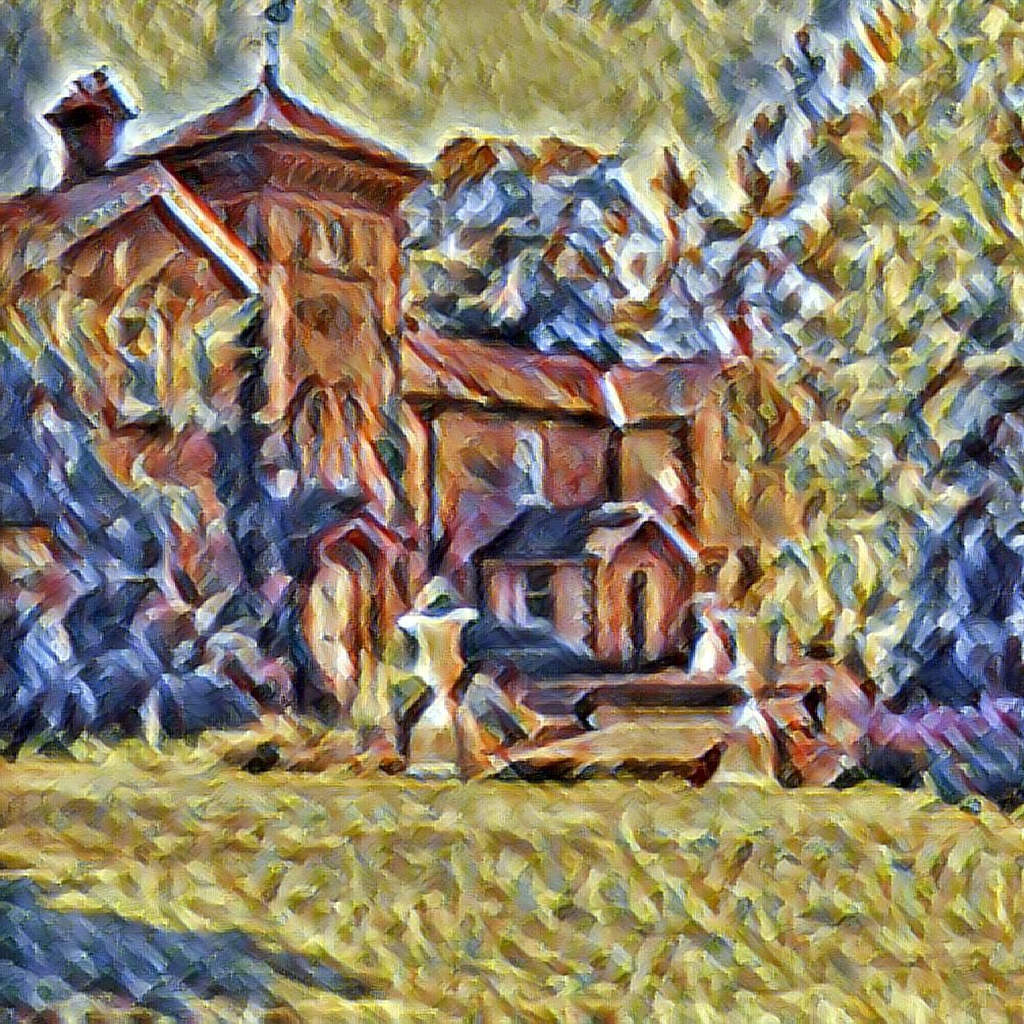}}
    \\ [-1ex]
    \subfloat[(a) Style]{\includegraphics[width = 1.7in,height = 1.7in]{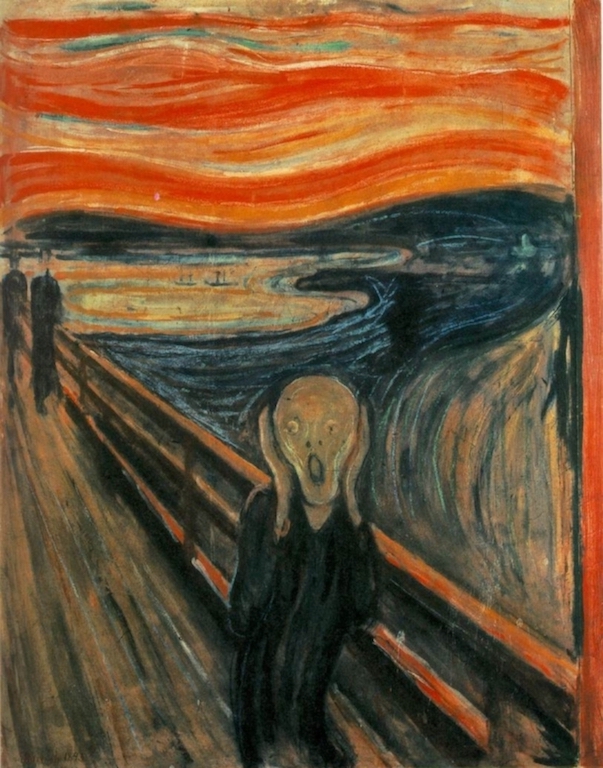}} & 
    \subfloat[(b) Singular Transfer (style size 256)]{\includegraphics[width = 1.7in]{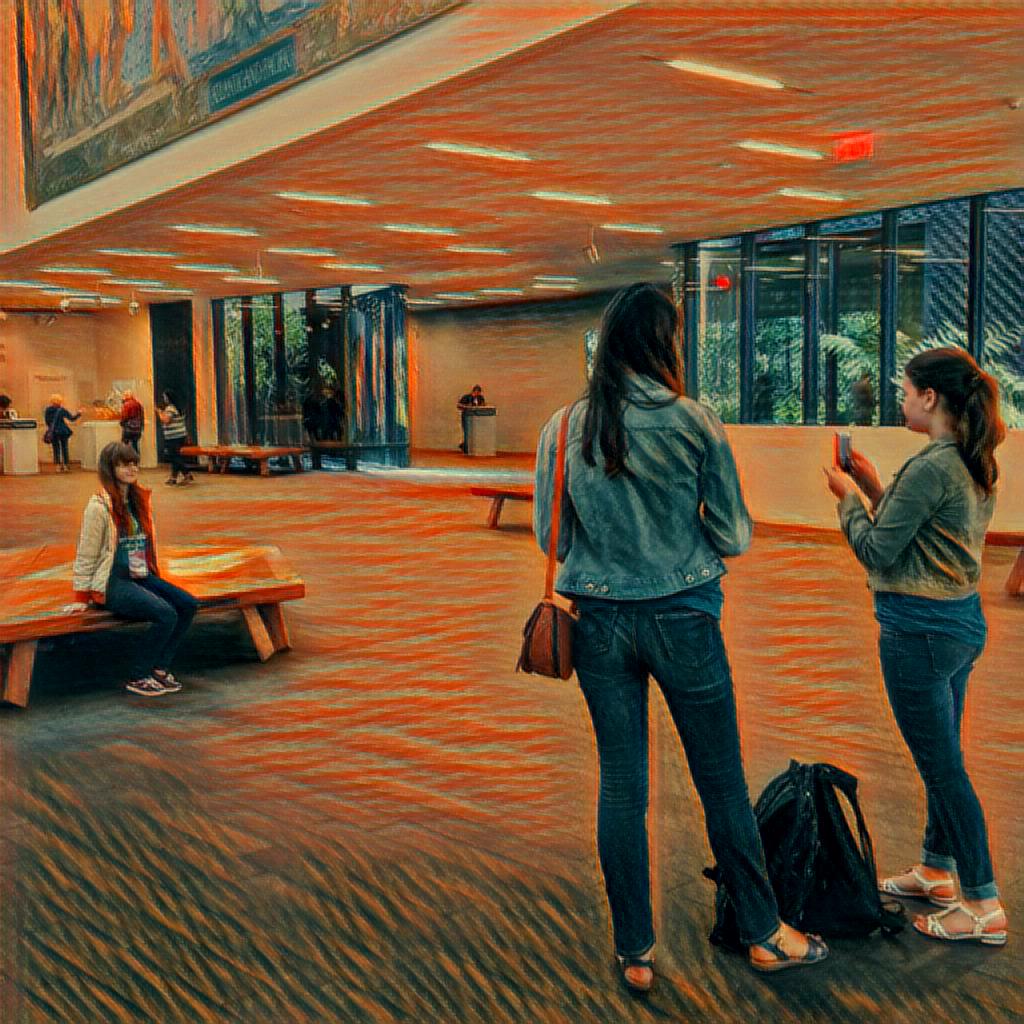}} & 
    \subfloat[(c) Singular Transfer (style size 1024)]{\includegraphics[width = 1.7in]{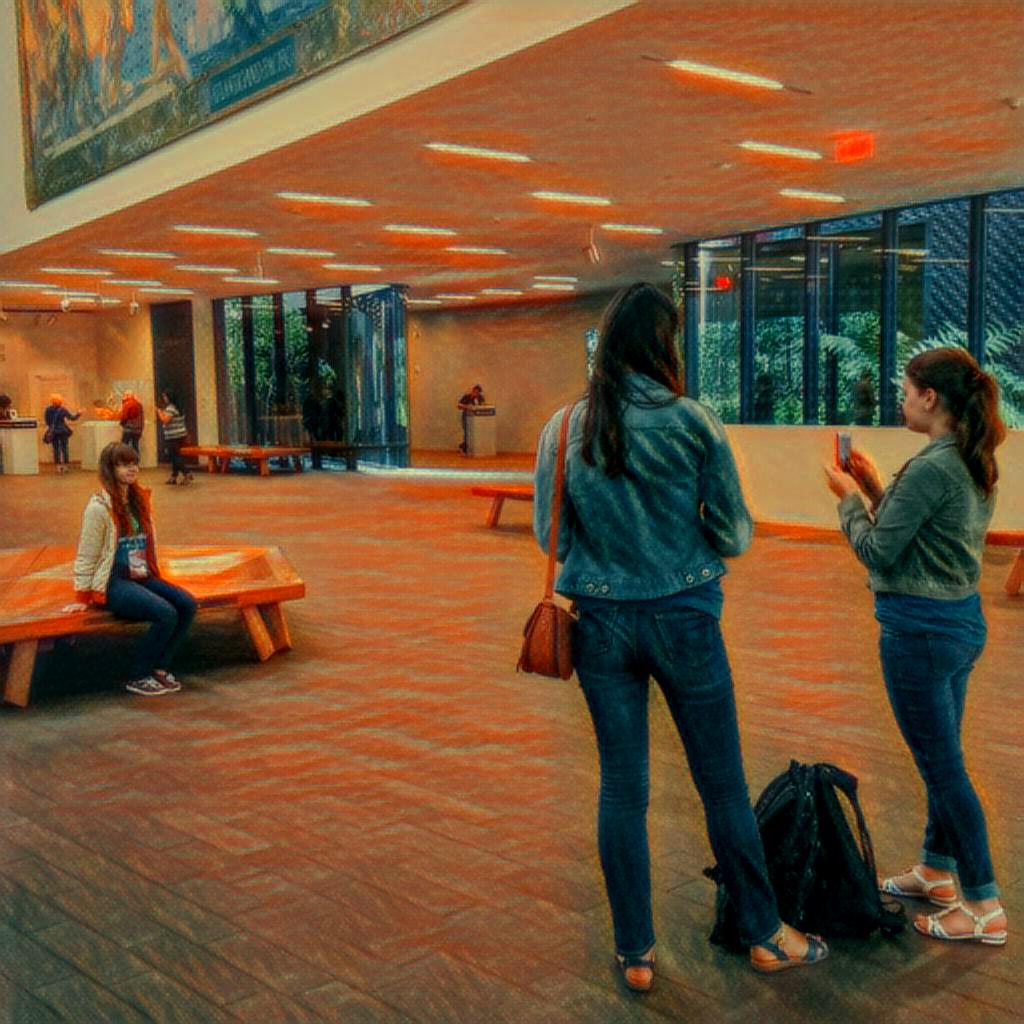}} &
    \subfloat[(d) Multimodal Transfer]{\includegraphics[width = 1.7in]{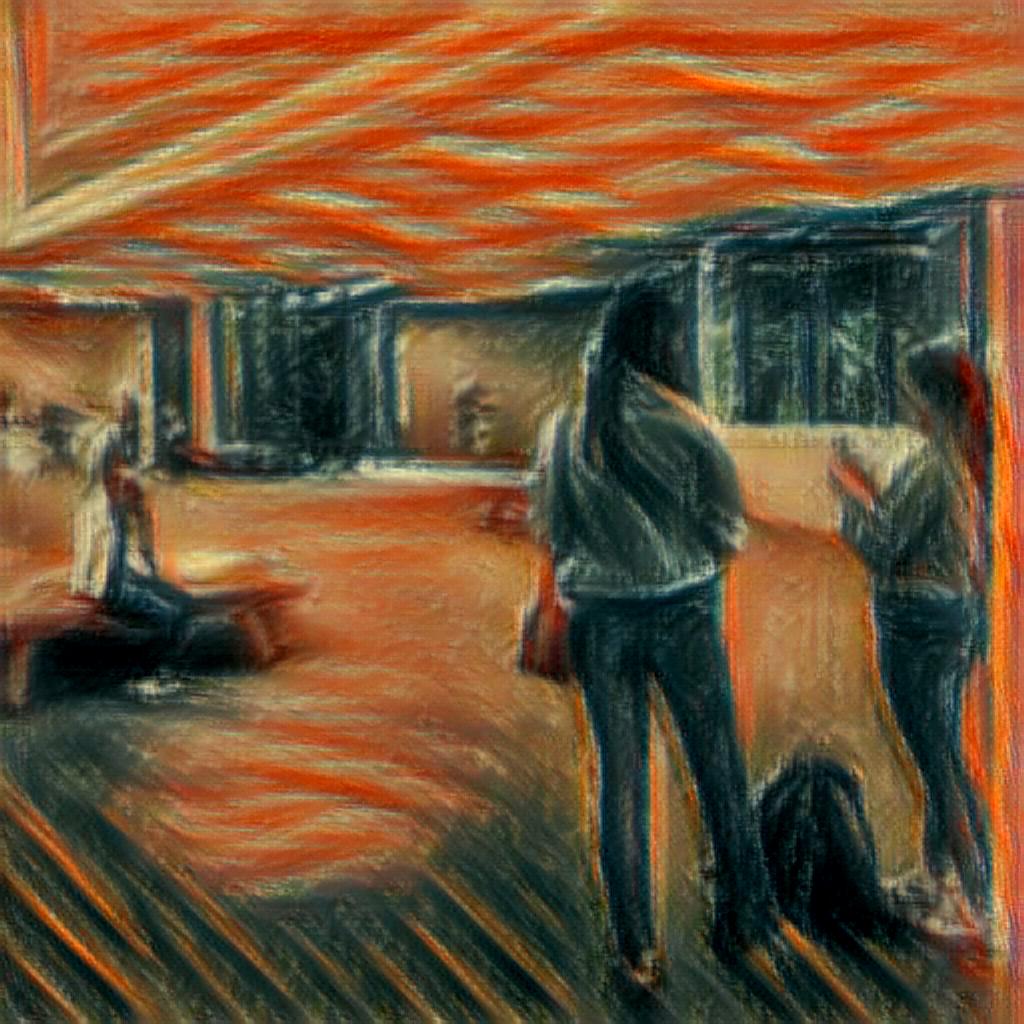}}
    \end{tabular}
\end{center}
   \caption{We compared our multimodal transfer with two singular transfer networks trained with different scales of the style image, 256 and 1024 (the singular transfer networks had the same architecture and had the same number of parameters as  our multimodal transfer network). All generated results were $1024 \times 1024$ pixels. As can be seen, there was  a big texture scale mismatch issue in the results of singular transfer 256 (column (b)) that the texture scale  was apparently smaller than that of the original artworks. While singular transfer 1024 (column (c)) failed to learn the texture distortion and brushwork although being rendering with the correct color. The results of multimodal transfer (column (d)) resolved those issues and managed to learn both coarse texture and fine, intricate brushwork.}

\label{fig:multimodal}
\end{figure*}

\subsubsection{Style Subnet} \label{sec:style}

\paragraph {Luminance-Color Joint Learning}
To better address the issue of preserving small intricate textures, our network utilizes representations of both color and luminance channels, because visual perception is far more sensitive to changes in luminance than in color~\cite{wandell1995foundations,hertzmann2001image,gatys2016preserving}. 
We separate the luminance channel from the RGB color image and use two independent branches (\textit{RGB-Block} and \textit{L-Block}) to learn their representations distinctively. 
The feature maps calculated from both branches are then joined together along the depth dimension and further processed by the ensuing \textit{Conv-Block}. 

\textit{RGB-Block} comprises three strided convolutional layers ($9 \times 9, 3 \times 3, 3 \times 3$ respectively, the latter two are used for downsampling) and three residual blocks \cite{he2016deep}, while \textit{L-Block} has a similar structure except that the depth of convolution is different. 
\textit{Conv-Block} is composed of three residual blocks, two resize-convolution layers for upsampling and the last $3 \times 3$ convolutional layer to obtain the output RGB image $\vec{\hat{y}}$. 
All non-residual convolutional layers are followed by instance normalization \cite{ulyanov2016instance} and ReLU nonlinearity. 
Part of our style subnet is designed based on the work \cite{johnson2016perceptual,radford2015unsupervised}.

A nearest neighbor interpolation upsampling layer and a  convolutional layer called \textit{resize-convolution layer} is  used here instead of deconvolutions to avoid the checkerboard artifacts of generated images~\cite{odena2016deconvolution}. 

\subsubsection{Enhance Subnet and Refine Subnet} \label{sec:enhance}

Although the style subnet is intended to stylize the input image with large texture distortion to match that of the style guide, we have found  that it is difficult to optimally adjust texture and content weights to achieve style transfer while preserving the content for a large variety of styles. 
Thus, we allow the style subnet to perform texture mapping with an eye toward preserving the content, and train a separate enhance subnet with a large texture weight to further enhance the stylization. 
Fig.~\ref{fig:diff} illustrates the specific role of each subnet. 
Evidently, the style subnet changes both color and texture heavily, but the enhance subnet also contributes greatly to the texture mapping while adding more detail. 
The refine net further refines and adds more detail into the final result. 

Accordingly, for the sake of enhancing the stylization, we adopt a similar structure as the style subnet for the enhance subnet. 
The only difference is that the enhance subnet has one more convolutional layer for downsampling and one more resize-convolution layer for upsampling, which enlarges the receptive field sizes. 
This is needed because the input to the enhance subnet is twice larger than that to the style subnet. 

Finally, the refine net consists of three convolutional layers, three residual blocks, two resize-convolution layers and one last convolutional layer to obtain the final output, which is much shallower than the style and enhance subnet. 
This is  because from Eq.~\eqref{eq:output} and \eqref{eq:min} we know former subnets can also contribute to the learning  tasks of the latter. 
Shortening the refine subnet is advantageous. 
It significantly reduces memory and computational complexity, which is critical for images of size 1024.
Furthermore, we add an identity connection from its beginning to the end, forcing it to learn just the difference between its input and output. 



\section{Experiments}
\label{sec:experiments}

\paragraph{Training Details}

The MT Network was trained on a subset of the Microsoft COCO dataset \cite{lin2014microsoft}, which contained 32,059 images (whose width and height were $\geq 480$). 
We cropped those images and resized them to $512 \times 512$. Adam optimization \cite{kingma2014adam} was used to train models for 10,000 iterations with batch size 1. 
The learning rate was initially set as $1 \times 10^{-3}$ and then reduced by a factor 0.8 every 2,000 iterations. 
Content losses were computed at layer relu4\_2 of VGG-19 and texture losses at layers relu1\_1, relu2\_1, relu3\_1 and relu4\_1 for all subnets. 
The content weights were all set to 1, while the texture weights depended on different styles, because a universal ratio of texture to content did not fit all artistic styles. 
As for the weights of stylization losses, we set $\lambda_1:\lambda_2:\lambda_3 = 1:0.5:0.25$. The rationale was that, during training, the parameters of former subnets were updated to incorporate the current and latter stylization losses. The latter losses should have smaller weights in order not to totally dominate the optimization process of the former subnets. Experiments revealed, however, that results were fairly robust to changes in $\lambda_k$. 
It took around one hour to fully train a hierarchical model on an NVIDIA GTX 1080. The model size on disk was about 35MB. 

\def\imagetop#1{\vtop{\null\hbox{#1}}}
\begin{figure}
\vspace*{-1ex}
\begin{center}

\subfloat[]{\includegraphics[width=1in,height=1in]{images/styles/picasso.jpg}}
~
\subfloat[]{\includegraphics[width=1in,height=1in]{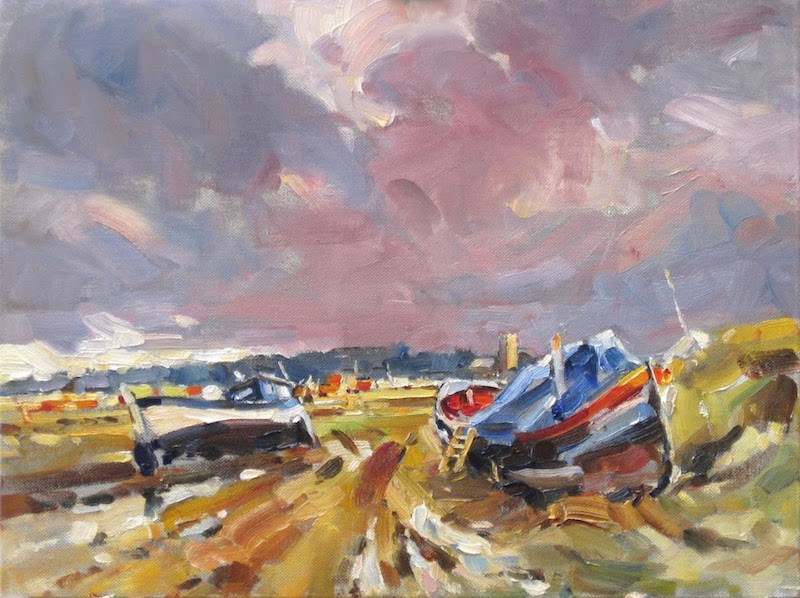}}
~
\subfloat[]{\includegraphics[width=1in]{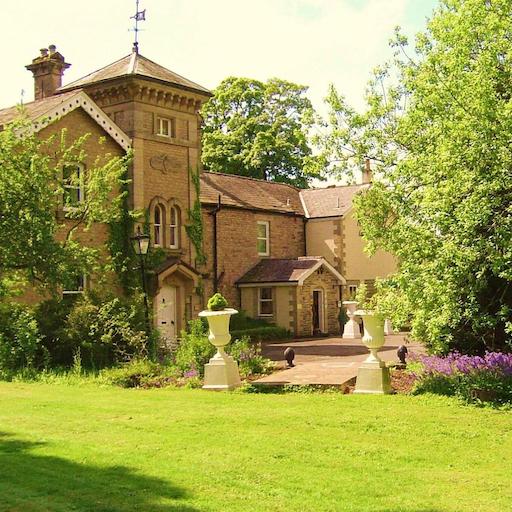}}
\\ \vspace{-2ex}
\subfloat[]{\includegraphics[width = 1in]{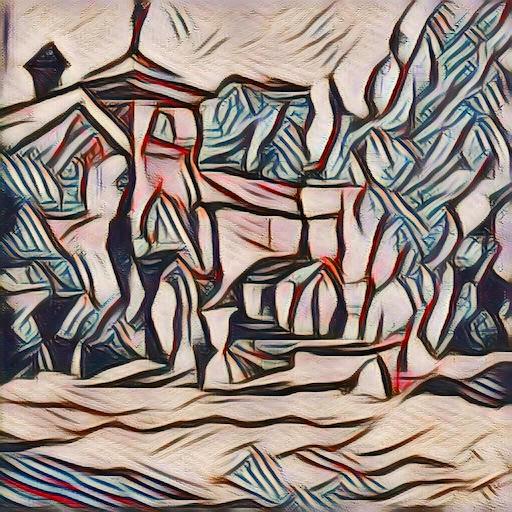}}
~
\subfloat[]{\includegraphics[width = 1in]{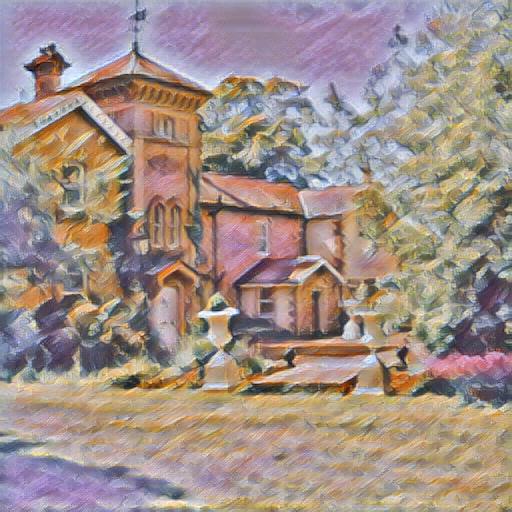}}
~
\subfloat[]{\includegraphics[width = 1in]{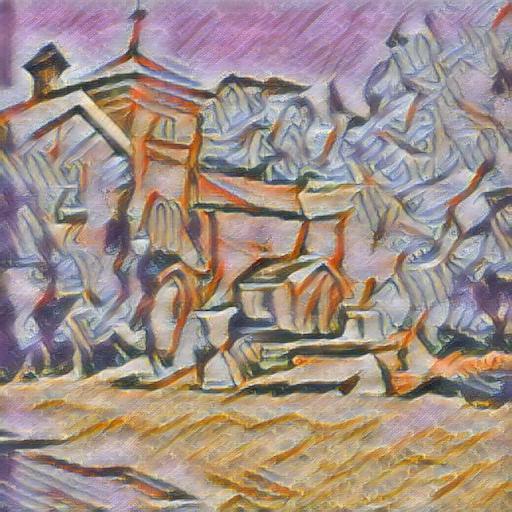}}
\vspace{-3ex}

\end{center}

   \caption{ Multimodal transfer with two styles. The model was trained with style (a) \textit{Still Life with Skull, Leeks and Pitcher} by Pablo Picasso, and (b) \textit{At the Close of Day} by Tomas King. (c) was the test image, and (f) was the final stylized result that had large texture distortion from (a) and small, detailed brushwork from (b). For comparison, we gave the  results transferred by the models trained on a single style in (d) and (e). }
\label{fig:multistyle}
\vspace*{-1ex}
\end{figure}

\paragraph{Comparison among Different Singular Transfer Networks}

As mentioned before, singular transfer was a feed-forward style-transfer network with a single stylization loss and multimodal transfer was the hierarchical network with multiple stylization losses and a mixture of modallities (color and luminance). 
Here we separated the style subnet from our MT network as a singular transfer network and compared it with the other state-of-the-art networks by Johnson \textit{ et al.}~\cite{johnson2016perceptual} and Ulyanov \textit{et al.}~\cite{ulyanov2016instance}. 
All three networks were trained on images of size 256 with the same content to texture weights and used instance normalization. Generally speaking, our style subnet generated qualitatively comparable results.
Particularly, it performed better than others on capturing texture details in some cases. 
In Fig.~\ref{fig:singular} we gave two comparison examples. 
In the first example, the result of Ulyanov \textit{et al.} was visibly darker than the  style image in color, and the texture scale in Johnson \textit{et al}'s result did not match that of the style image very well. 
The result of our style subnet seemed better in both aspects. 
In the second example, comparing with the other two networks, our style subnet performed better on simulating small, detailed texture. 
Therefore, we chose the style subnet as a representative of singular transfer to be compared with multimodal transfer next. 

\paragraph{Singular Transfer \textit{VS} Multimodal Transfer on High-resolution Images}  

We tested our method on numerous artistic styles. In Fig.~\ref{fig:multimodal}, we compared our multimodal transfer network on high-resolution images ($1024 \times 1024$) with a singular transfer network with the same number of learning weights. 
(More exactly, we duplicated our style subnet to a deeper network that had the same number of parameters to provide a fair comparison). 
Examining the results shown in Fig.~\ref{fig:multimodal}, comparing with singular transfer, multimodal transfer results were visually more similar to the original artistic styles both in coarse texture structure and fine brushwork, while singular transfer with the style size 256 caused a texture scale mismatch that the texture scale was much smaller than that of the original artwork. 
Furthermore, singular transfer with style size 1024 failed to learn the distortion and fine brushwork.

\paragraph{Multimodal Transfer with Multiple Styles}

Our multimodal transfer allowed an interesting application that was not possible before: 
It could  be trained with multiple styles such that the final stylized result fused the content of one test image, the coarse texture distortion of one style image, and the fine brushwork of another style image. 
Here we gave an example in Fig.~\ref{fig:multistyle} where the model was trained with two different styles. 

\paragraph{Processing Speed and Memory Use}

We compared quantitatively the speed and memory usage of our multimodal transfer network (MT Net) with other singular transfer networks (Here we used Johnson Net, the network by Johnson \textit{et al.}).
We also constructed a deep singular transfer network for comparison (called DS Net), which had the same structure as the MT Net. 
We took the average of the test time for 1,000 generations (excluding model loading time). 

\begin{table}[!h]
\renewcommand{\arraystretch}{0}
\begin{center}
  \begin{tabular}{ |c||c|c|  }
   \hline
   Network & Test Time & Memory Usage \\
   \hline\hline
   MT Net &  0.54s  & ~3100 MB  \\
   \hline
   Johnson Net &  0.42s  & ~2400 MB  \\
   \hline
   DS Net &  0.63s  & ~6700 MB  \\
   \hline
  \end{tabular}
\end{center}
\vspace{-2ex}
\caption{Comparison results of speed and memory use when applied on $1024 \times 1024$ images.}
\vspace{-1ex}
\label{table:time}
\end{table}

As shown in Table \ref{table:time}, although MT Net was more than twice deeper than Johnson Net, its speed and memory usage were close to those of Johnson Net (0.54s \textit{vs} 0.42s, 3100 MB \textit{vs} 2400 MB) when generating high-resolution images, which benefited from the hierarchical transfer procedure where most computation was done on low resolutions.  
The singular transfer network DS Net performed the worst even with same number of parameters. 
Therefore, multimodal transfer is suitable for real-world applications that usually required high image resolution, because it was able to generate results more similar to the desired artistic styles with a small cost.    

\section{Conclusion}
\label{sec:conclusion}

Here we present a hierarchical training scheme (\textit{multimodal transfer}) for fast style transfer to learn artistic style cues at multiple scales, including color, coarse texture structure and fine, exquisite brushwork. 
The scheme solves the texture scale mismatch issue and generates much more visually appealing stylized results on high-resolution images. 

In the future, we plan to investigate other losses that can better capture  the artistic style at different scales. 
We also want to explore alternate loss networks that costs less memory to extend our scheme onto much larger images.

{\small
\bibliographystyle{ieee}
\bibliography{egbib}
}

\end{document}